\documentclass[11pt]{article}

\usepackage{fullpage}
\usepackage{mathptmx}
\usepackage{times}
\usepackage{microtype}
\usepackage{graphicx}
\usepackage{subfigure}
\usepackage{booktabs} 
\usepackage{amsmath,amsbsy,amsfonts,amssymb,amsthm}
\usepackage{color}
\usepackage{algorithmic,mathtools}
\usepackage{algorithm}
\usepackage[utf8]{inputenc} 
\usepackage[T1]{fontenc}    
\usepackage{hyperref}       
\usepackage{url}            
\usepackage{booktabs}       
\usepackage{nicefrac}       
\usepackage{microtype}      
\usepackage{xcolor}         

\usepackage{xspace}
\usepackage{thm-restate}

\usepackage{wrapfig}

\newcommand{\alert}[1]{{\textcolor{red}{#1}}}
\newcommand{\cA}{{\mathcal A}}
\newcommand{\dbl}{{\sc Double}\xspace}

\newcommand{\pdbl}{{\sc Predict-and-Double}\xspace}

\newcommand{\pdim}{{\sc Pdim}\xspace}

\newcommand{\eat}[1]{}

\DeclarePairedDelimiter\abs{\lvert}{\rvert}%
\DeclarePairedDelimiter\norm{\lVert}{\rVert}%

\makeatletter
\let\oldabs\abs
\def\abs{\@ifstar{\oldabs}{\oldabs*}}
\let\oldnorm\norm
\def\norm{\@ifstar{\oldnorm}{\oldnorm*}}
\makeatother

\DeclareMathOperator*{\argmin}{arg\,min}

\newcommand{\comp}{\textsc{cr}\xspace}

\newcommand{\Off}{{\textsc{opt}\xspace}}
\newcommand{\cX}{{\mathbb{X}}}
\newcommand{\cY}{{\mathbb{Y}}}
\newcommand{\ex}{\mathbb{E}}
\newcommand{\cF}{{\mathcal{F}}}

\newcommand{\cP}{{\cal P}}
\newcommand{\cI}{{\cal I}}
\newcommand{\dist}{{\mathbb{D}}}
\newcommand{\err}{\textbf{\textsc{er}}\xspace}
\newcommand{\Ast}{{\mathcal A}^\star}

\newtheorem{theorem}{Theorem}
\newtheorem{definition}{Definition}
\newtheorem{corollary}[theorem]{Corollary}
\newtheorem{lemma}[theorem]{Lemma}

\newcommand{\OS}{{\textsc{OnlineSearch}}\xspace}
\newcommand{\ltos}{{\textsc{LearnToSearch}}\xspace}
\newcommand{\opt}{{\textsc{opt}}\xspace}
\newcommand{\sol}{{\textsc{sol}}\xspace}
\newcommand{\minlength}{{\textsc{Min-Length}}\xspace}
\newcommand{\maxlength}{{\textsc{Max-Length}}\xspace}

\newcommand{\ml}{{\sc ml}\xspace}

\PassOptionsToPackage{numbers}{natbib}

\title{Customizing ML Predictions for Online Algorithms}
\author{Keerti Anand\thanks{Department of Computer Science, Duke University, Durham, NC, USA. Emails: {\tt \{kanand, rongge, debmalya\}@cs.duke.edu}.}
\and Rong Ge\footnotemark[1]
\and Amit Kumar \thanks{Department of Computer Science, IIT Delhi, New Delhi, India. Email: {\tt amitk@cse.iitd.ernet.in} }\footnotemark[2]
\and Debmalya Panigrahi\footnotemark[1]
}

\date{}

\title{A Regression Approach to Learning-Augmented Online Algorithms}

\begin{document}

\maketitle

\begin{abstract}
The emerging field of learning-augmented online algorithms uses ML techniques to predict future input parameters and thereby improve the performance of online algorithms. Since these parameters are, in general, real-valued functions, a natural approach is to use regression techniques to make these predictions. We introduce this approach in this paper, and explore it in the context of a general online search framework that captures classic problems like (generalized) ski rental, bin packing, minimum makespan scheduling, etc. We show nearly tight bounds on the sample complexity of this regression problem, and extend our results to the agnostic setting. From a technical standpoint, we show that the key is to incorporate online optimization benchmarks in the design of the loss function for the regression problem, thereby diverging from the use of off-the-shelf regression tools with standard bounds on statistical error.
\end{abstract}

\clearpage

\section{Introduction}
\label{sec:intro}
A recent trend in {\em online algorithms} has seen the use of {\em future predictions} generated by ML techniques to bypass pessimistic worst-case lower bounds. A growing body of work has started to emerge in this area in the last few years addressing a broad variety of problems in online algorithms such as rent or buy, caching, metrical task systems, matching, scheduling, experts learning, stopping problems, and others (see related work for references).
The vast majority of this literature is focused on {\em using} ML predictions in online algorithms, but does not address the question of how these predictions are generated. This raises the question: {\em what can we learn from data that will improve the performance of online algorithms?} Abstractly, this question comes in two inter-dependent parts: the first part is a {\em learning} problem where we seek to learn a function that maps the feature domain to predicted parameters, and the second part is to re-design the online algorithm to use these predictions. In this paper, we focus on the first part of this design pipeline, namely {\em we develop a regression approach to generating ML predictions for online algorithms}.

Before delving into this question further, we note that there has been some recent research that focuses on the learnability of predicted parameters in online algorithms. Recently, Lavastida~{\em et al.}~\cite{lavastida2020learnable}, building on the work of Lattanzi~{\em et al.}~\cite{LattanziLMV20}, took a data-driven algorithms approach to design online algorithms for scheduling and matching problems via {\em learned weights}. In this line of work, the goal is to observe sample inputs in order to learn a set of weights that facilitate better algorithms for instances from a fixed distribution. In contrast, Anand {\em et al.}~\cite{AnandGP20} relied on a classification learning approach for the Ski Rental problem, where they aimed to learn a function that maps the feature set to a binary label characterizing the optimal solution. But, in  general, the value of the optimal solution is a real-valued function, which motivates a regression approach to learning-augmented online algorithms that we develop in this paper.

To formalize the notion of an unknown optimal solution that we seek to learn via regression, we use the {\em online search} (\OS) framework. 
In this framework, there is as an input sequence $\Sigma = \sigma_1, \sigma_2, \ldots$ available offline, and the actual online input is a prefix of this sequence $\Sigma_T = \sigma_1, \sigma_2, \ldots, \sigma_T$, where the length of the prefix $T$ is revealed online. Namely, in each online step $t > 0$, there are two possibilities: either the sequence {\em ends}, i.e., $T = t$, or the sequence {\em continues}, i.e., $T > t$. The algorithm must maintain, at all times $t$, a solution that is feasible for the current sequence, i.e., for the prefix $\Sigma_t = \sigma_1, \ldots, \sigma_t$. The goal is to obtain a solution that is of minimum cost among all the feasible solutions for the actual input sequence $\Sigma_T$.

We will discuss applicability of the \OS framework in more detail in Section~\ref{sec:applicability}, but for a quick illustration now, consider the ski rental problem in this framework. In this problem, if the sequence continues on day $t$, then the algorithm must rent skis if it has not already bought them. In generalizations of the ski rental problem to multiple rental options, the requirement is that one of the rental options availed by the algorithm must cover day $t$. We will show in Section~\ref{sec:applicability} that we can similarly model several other classic online problems in the \OS framework.


We use the standard notion of {\em competitive ratio}, defined as the worst case ratio between the algorithm's cost and the off-line optimal cost, to quantify the performance of an online algorithm. For online algorithms with predictions, we follow the terminology in \cite{PurohitSK18} that is now standard: we say that the {\em consistency} and {\em robustness} of an algorithm are its competitive ratios for correct predictions and for arbitrarily incorrect predictions respectively. Typically, we fix consistency at $1+\epsilon$ for a hyper-parameter $\epsilon$ and aim to minimize robustness as a function of $\epsilon$.

We make some mild assumptions on the problem. First, we assume that solutions are {\em composable}, i.e., that adding feasible solutions for subsequences ensures feasibility over the entire sequence; second, that cost is {\em monotone}, i.e., the optimal cost for a subsequence is at most that for the entire sequence; and third, that the offline problem is (approximately or exactly) {\em solvable}. These assumptions hold for essentially all online problems we care for.

\subsection{Our Contributions}
\label{sec:contributions}
\eat{In this paper, we work in a learning framework where the optimal cost for a certain problem instance of optimal solution is determined by a certain set of features. Moreover, there exists a class of functions that map from the space of features to the cost of the optimal solution.}

As a warm up, we first give an algorithm called \dbl for the \OS problem {\em without predictions} in Section~\ref{sec:prelim}. The \dbl algorithm has a competitive ratio of $4$. We build on the \dbl algorithm in Section~\ref{sec:blackbox}, where we give an algorithm called \pdbl for the \OS problem {\em with predictions}. We show that the \pdbl algorithm has a consistency of $1+\epsilon$ and robustness of $O(1/\epsilon)$, for any hyper-parameter $\epsilon > 0$. We also show that this tradeoff between consistency and robustness is asymptotically tight.

Our main contributions are in Section~\ref{sec:whitebox}. In this section, we model the question of obtaining a learning-augmented algorithm for the \OS problem in a regression framework. Specifically, we assume that the input comprises a feature vector $x$ that is mapped by an unknown real-valued function $f$ to an input for the \OS problem $z$. In the training phase, we are given a set of labeled samples of the form $(x, z)$ from some (unknown to the algorithm) data distribution $\mathbb{D}$. The goal of the learning algorithm is to produce a mapping from the feature space to algorithmic strategies for the \OS problem, such that when it gets an unlabeled (test) sample $x$ from the same distribution $\mathbb{D}$, the algorithmic strategy corresponding to $x$ obtains a competitive solution for the actual input $z$ in the test sample (that is unknown to the algorithm).

The learning algorithm employs a regression approach in the following manner. It assumes that the function $f$ is from a hypothesis class $\mathcal{F}$, and obtains an empirical minimizer in $\mathcal{F}$ for a carefully crafted loss function on the training samples. The design of this loss function is crucial since a bound on this loss function is then shown to translate to a bound on the competitive ratio of the algorithmic strategy. (Indeed, we will show later that because of this reason, standard loss functions used in regression are inadequate for our purpose.) Finally, we use statistical learning theory for real-valued functions to bound the sample complexity of the learner that we design. 

Using the above framework, we show a sample complexity bound of $O\left(\frac{H\cdot d}{\epsilon}\right)$ for obtaining a competitive ratio of $1+\epsilon$, where $H$ and $d$ respectively represent the log-range of the optimal cost and a measure of the expressiveness of the function class $\mathcal{F}$ called its pseudo-dimension.\footnote{Intuitively, the notion of pseudo-dimension extends that of the well-known VC dimension from binary to real-valued functions.} We also extend this result to the so-called agnostic setting, where the function class $\mathcal{F}$ is no longer guaranteed to contain an exact function $f$ that maps $x$ to $z$, rather the competitive ratio is now in terms of the {\em best} function in this class that approximates $f$. We also prove nearly matching lower bounds for our sample complexity bounds in the two models.

Our framework can also be extended to the setting where the offline optimal solution is hard to compute, but there exists an algorithm with competitive ratio $c$ given the {\em cost} of optimal solution. In that case our algorithms gives a competitive ratio $c(1+\epsilon)$, which can still be better than the competitive ratio without predictions (see examples in next subsection).
\eat{
This is matched (ignoring log factors) by a corresponding lower bound.
When such a determination is not perfect, we define a benchmark that captures how well the class of functions predict the behavior of the cost of the optimal solution with respect to the features. This entails defining our own loss metric which we term as ``competitive loss'', that we show is more suitable for evaluating our predictions. Given, that the best predictor in our class has competitive error $\Delta$, we design an algorithm that has competitive ratio $1+O(\Delta)$ that uses $O(\frac{1}{\Delta})$ samples. On the other hand, we show that in the general case, to get $1+\epsilon$ factor close to the optimal algorithm, one requires $\Omega(\frac{1}{\epsilon^2})$ samples. 
}


\subsection{Applicability of the \OS framework}
\label{sec:applicability}

The \OS framework is applicable {\em whenever an online algorithm benefits from knowing the optimal value of the solution}. Many online problems benefit from this knowledge, which is sometimes called {\em advice} in the online algorithms literature. For concreteness, we give three examples of classic problems -- {\em ski rental with multiple options}, {\em online scheduling}, and {\em online bin packing} -- to illustrate the applicability of our framework. Our algorithm PREDICT-AND-DOUBLE (explained in more detail in section~\ref{sec:blackbox}) successively predicts the optimal value of the solution and appends the corresponding solution to its output.

\textbf{Ski Rental with Multiple Options.}
Generalizations of the ski rental problem with multiple options have been widely studied (e.g., \cite{AiWHHTL14,LotkerPR12,Meyerson05,Fleischer01}), recently with \ml predictions~\cite{WangLW20}.
Suppose there are $V$ options (say coupons) at our disposal, where coupon $i$ costs us $C_i$ and is valid for $d_i$ number of days. Given such a setup, we need to come up with a schedule: $\{(t_{k}, i_{k}), k=1,2\ldots \}$ that instructs us to buy coupon $i_k$ at time $t_k$. (The classic ski rental problem corresponds to having only two coupons $C_1=1, d_1 =1$ and $C_2 = B, d_2 \rightarrow \infty$.) Our $\OS$ framework is applicable here: 
a solution that allows us to buy  coupons valid time $t$ is also a valid solution for all times $s\le t$. Further, PREDICT-AND-DOUBLE can be implemented efficiently as we can compute $\opt(t)$, for any time $t$ using a dynamic program. 


\textbf{Online Scheduling.}  
Next, we consider the classic online scheduling problem where the goal is to assign jobs arriving online to a set of identical machines so as to minimize the maximum load on any machine (called the {\em makespan}). 
For this algorithm, the classic list scheduling algorithm~\cite{Graham69} has a competitive ratio of $2$. 
A series of works \cite{galambos1993line, BartalFKV92, karger1996better, albers1999better} improved the competitive ratio to 1.924, and currently the best known result has competitive ratio of (approx) $1.92$~\cite{FleischerW00}; in fact, there are nearly matching lower bounds \cite{gormley2000generating}. However, if the optimal makespan ($\opt$) is {\em known}, then these lower bounds can be overcome, and a significantly better competitive ratio of $1.5$ can be obtained in this setting~\cite{BohmSSV17} (see also \cite{DBLP:conf/random/AzarR98,KellererK13,GabayKB15,GabayBK17}). The \OS framework is applicable here with a slight modification: whenever PREDICT-AND-DOUBLE tries to buy a solution corresponding to a predicted value of $\opt$, we execute the 1.5-approximation algorithm based on this value. The problem still satisfies the property that a solution for $t$ jobs is valid for any prefix. We get a competitive ratio of $1.5+O(\epsilon)$ that significantly outperforms the competitive ratio of $1.92$ without predictions.

\textbf{Online Bin Packing.}
As a third example, we consider the online bin packing problem. In this problem, items arrive online and must be packed into fixed-sized bins, the goal being to minimize the number of bins. (We can assume w.l.o.g., by scaling, that the bins are of unit size.) Here, it is known that the critical parameter that one needs to know/predict is not $\opt$ but the number of items of {\em moderate} size, namely those sized between $1/2$ and $2/3$. If this is known, then there is a simple $1.5$-competitive algorithm~\cite{AngelopoulosDKR15}, which is not achievable without this additional knowledge. Again, our \OS framework can be used to take advantage of this result. In this case, the application is not as direct, because predicting $\opt$ does not yield the better algorithm. Nevertheless, an inspection of the algorithm in~\cite{AngelopoulosDKR15} reveals the following strategy: The items are partitioned into three groups. The items of size $\ge 2/3$ are assigned individual bins, items of size between $1/3$ and $1/2$ are assigned separate bins where at least two of them are assigned to each bin, and the remaining items are assigned a set of common bins. Clearly, the first two categories can be handled online without any additional information; this means that we can define a surrogate $\opt$ (call it $\opt'$) that only captures the optimal number of bins for the common category. Note that the of prediction of OPT’ serves as a substitute for knowing the numbers of items of moderate size. Now, if $\opt'$ is known, then we can recover the competitive ratio of $3/2$ by using a simple greedy strategy. This now allows us to use the \OS framework where we predict $\opt'$. As earlier, the \OS framework can be applied with slight modification: whenever PREDICT-AND-DOUBLE tries to buy a solution corresponding to a predicted value of $\opt'$, we execute the 1.5-competitive algorithm based on this value. The problem still satisfies the property that a solution for $t$ items is valid for any prefix.

\subsection{Motivation for a cognizant loss function}
\label{sec:motivation}
In this work, we explore the idea of a carefully crafted loss function that can help in making better predictions for the online decision task. To illustrate this, consider the problem of balancing the load between machines/clusters in a data center where remote users are submitting jobs. The goal is to minimize the maximum load on any machine, also called the makespan of the assignment. The optimal makespan, which we would like to predict, depends on the workload submitted by individual users who are currently active in the system. Therefore, we would like to use the user features to predict their behavior in terms of the workload submitted to the server. A typical feature vector would then be a binary vector encoding  the set of users currently active in the system, and based on this information, a learning model trained on historical behavior of the users can predict (say) a histogram of loads that these users are expected to submit, and therefore, the value of the optimal makespan. The feature space can be richer, e.g., including contextual information like the time of the day, day of the week, etc. that are useful to more accurately predict user behavior. Irrespective of the precise learning model, the main idea in this paper is that the learner should try to optimize for competitive loss instead of standard loss functions. This is because the goal of the learner is not to accurately predict the workload of each user, but to eventually obtain the best possible makespan. For instance, a user who submits few jobs that are inconsequential to the eventual makespan need not be accurately predicted. Our technique automatically makes this adjustment in the loss function, thereby obtaining better performance on the competitive ratio.
\subsection{Related Work}
There has been considerable recent work in incorporating \ml predictions in online algorithms. Some of the problems include: auction pricing~\cite{MedinaV17}, ski rental~\cite{PurohitSK18,GollapudiP19,AnandGP20,Banerjee20,WangLW20,AngelopoulosDJKR20}, caching ~\cite{LykourisV18,Rohatgi20,JiangPS20, Wei20}, scheduling~\cite{PurohitSK18,LattanziLMV20,Mitzenmacher20}, frequency estimation~\cite{HsuIKV19}, Bloom filters~\cite{Mitzenmacher18}, online linear optimization~\cite{bhaskara2020online}, speed scaling~\cite{BamasMRS20}, set cover~\cite{BamasMS20}, bipartite and secretary problems~\cite{AntoniadisGKK20}, etc. 
While most of these papers focus on designing online algorithms for ML predictions but not on the generation of these predictions, there has also been some work on the design of predictors using a binary classification approach~\cite{AnandGP20}), and on the formal learnability of the predicted parameters \cite{LattanziLMV20,lavastida2020learnable}. In contrast, we use a regression approach to the problem in this paper.

The PAC learning framework was first introduced by \cite{valiant1984theory} in the context of learning binary classification functions, and related the sample complexity to the VC dimension of the hypothesis class. This was later extended to real-valued functions by \cite{pollard1990empirical}, who introduced the concept of pseudo-dimension, and \cite{KearnsS94} (see also \cite{BartlettLW96}), who introduced the fat shattering dimension, as generalizations of VC dimension to real-valued functions. For a comprehensive discussion of the extension of VC theory to learning real-valued functions, the reader is referred to the excellent text by \cite{anthony1999learning}. A different approach was proposed by \cite{alon1997scale} (see also \cite{DBLP:journals/cpc/AnthonyB00}) who analysed a model of learning in which the error of a hypothesis is taken to be the expected squared loss, and gave uniform convergence results for this setting. 
In this paper, we use pseudo-dimension and corresponding sampling complexity bounds in quantifying the complexity of the regression learning problem of predicting input length.


\section{{\sc OnlineSearch} without Predictions}
\label{sec:prelim}
As a warm up, we first describe a simple algorithm called \dbl (Algorithm~\ref{Alg: double}) for the \OS problem {\em without predictions.} This algorithm places milestones on the input sequence corresponding to inputs at which the cost of the optimal solution doubles. When the input sequence crosses such a milestone, the algorithm buys the corresponding optimal solution and adds it to the existing online solution. This simple algorithm will form a building block for the algorithms that we will develop later in the paper; hence, we describe it and prove its properties below.
First, we introduce some notation.
\begin{definition}
We use $\Off(t)$ to denote an optimal (offline) solution for the input prefix of length $t$; we overload notation to denote the cost of this solution by $\Off(t)$ as well.
\end{definition}

\begin{definition}
Given an input length $\tau$ and any $\alpha > 0$, we use $\minlength(\alpha, \tau)$ to denote the smallest length $t$ such that $\Off(t) \geq \alpha \cdot  \Off(\tau)$.
The monotonicity property of $\Off$ ensures that $\minlength(\alpha, \tau) > \tau$ if $\alpha > 1$, and $\minlength(\alpha, \tau) \le \tau$ otherwise. 
\end{definition}

\eat{
\begin{algorithm}
\caption{\dbl}
\label{Alg: double}
\begin{algorithmic}
 \STATE {\bf Input:} The input sequence $\cI$.
 \STATE {\bf Output:} The online solution $\sol$.
 \STATE
 \STATE Set $i:=0$, $\tau_0:= 1$, $\sol := \emptyset$.
\FOR {$t=1,2, \ldots, T$}
      \IF{$t=\tau_i$}
            \STATE Set $\tau_{i+1} = \minlength(2, \tau_i)$.
            \STATE Add $\Off(\tau_{i+1}-1)$ to $\sol$.
            \STATE Increment $i$.
      \ENDIF
\ENDFOR
\end{algorithmic}
\end{algorithm}
}

\begin{algorithm}
\caption{\dbl}
\label{Alg: double}
 {\bf Input:} The input sequence $\cI$. \\
 {\bf Output:} The online solution $\sol$. \\
 
 Set $i:=0$, $\tau_0:= 1$, $\sol := \emptyset$. \\
    \textbf{for} $t=1,2, \ldots, T$ \\
        \hspace*{10pt} \textbf{if} $t=\tau_i$ \\
        \hspace*{10pt}\hspace*{10pt} Set $\tau_{i+1} = \minlength(2, \tau_i)$. \\
        \hspace*{10pt}\hspace*{10pt} Add $\Off(\tau_{i+1}-1)$ to $\sol$.\\
        \hspace*{10pt}\hspace*{10pt} Increment $i$.
\end{algorithm}

\begin{theorem}
\label{thm:dbl}
    The \dbl algorithm is $4$-competitive for the \OS problem.
\end{theorem}
\begin{proof}
Recall that $T$ denotes the length of the input sequence. Let $\tau_{i} \le T < \tau_{i+1}$. Then, the cost of the optimal solution, $\Off(T) \ge \Off(\tau_{i})$ by monotonicity, while the cost of the online solution $\sol$ is given by:
\begin{align*}
\Off(\tau_{1}-1) + &\Off(\tau_{2}-1) + \ldots + \Off(\tau_{i+1}-1)\\
&\le 2 \cdot \Off(\tau_{i+1}-1) \le 4\cdot \Off(\tau_{i}).\qedhere    
\end{align*}
\end{proof}

\section{{\sc OnlineSearch} with Predictions}
\label{sec:blackbox}
In the previous section, we described a simple online algorithm for the \OS problem. Now, we build on this algorithm to take advantage of ML predictions. For now, we do not concern ourselves with how these predictions are generated; we will address this question in the next section.

Suppose we have a prediction $\hat{T}$ for the input length $T$ of an \OS problem instance. Na\"ively, we might trust this prediction completely and buy the solution $\Off(\hat{T})$. 
While this algorithm is perfect if the prediction is accurate, it can fail in two ways if the prediction is inaccurate: (a) if $T \ll \hat{T}$ and therefore $\opt(T) \ll \opt(\hat{T})$, then the algorithm has a large competitive ratio, and (b) if $T > \hat{T}$, then $\opt(\hat{T})$ may not even be feasible for $T$. A natural idea is to then progressively add $\Off(t)$ solutions for small values of $t$ (similar to \dbl) until a certain threshold is reached, before buying the predicted optimal solution $\opt(\hat{T})$. Next, if $T > \hat{T}$, the algorithm can resume buying solution $\opt(t)$ for $t > \hat{T}$, again using \dbl, until the actual input $T$ is reached. 

One problem with this strategy, however, is that the algorithm does not degrade gracefully around the prediction, a property that we will need later in the paper. In particular, if $T$ is only slighter larger than $\hat{T}$, then the algorithm adds a solution that has cost $2 \cdot \opt(\hat{T})$, thereby realizing the worst case scenario in Theorem~\ref{thm:dbl} that was achieved without any prediction.
%
%
Our work-around for this issue is to buy $\opt(t)$ for a $t$ slightly larger than $\hat{T}$, instead of $\opt(\hat{T})$ itself, which secures us against the possibility of the actual input being slightly longer than the prediction. We call this algorithm \pdbl (Algorithm~\ref{Alg: online_input_seq_with_thresh}). Here, we use a hyper-parameter $\epsilon$ that offers a tradeoff between the consistency and robustness of the algorithm. We also use the following definition:
\begin{definition}
Given an input length $\tau$ and any $\alpha > 0$, we use $\maxlength(\alpha, \tau)$ to denote the largest length $t$ such that $\Off(t) \le \alpha \cdot  \Off(\tau)$.
\end{definition}

\eat{
\begin{algorithm}
\caption{\pdbl}
\label{Alg: online_input_seq_with_thresh}
\begin{algorithmic}
 \STATE {\bf Input:} The input sequence $\cI$ and prediction $\hat{T}$. 
 \STATE {\bf Output:} The online solution $\sol$.
 \STATE
 \STATE Set $\sol:= \emptyset$, 
        $t_1:= \minlength(\epsilon/5, \hat{T})$, and 
        $t_2:= \maxlength(1+\epsilon/5, \hat{T})$.
 \STATE {\bf Phase 1:} Execute \dbl while $t < t_1$.
 \STATE {\bf Phase 2:} At $t = t_1$, add $\Off(t_2)$ to $\sol$.
 \STATE {\bf Phase 3:} If $t > t_2$, resume \dbl as follows:
 \STATE Set $i:=0$, $\tau_0:= t_2+1$.
 \FOR {$t=t_2+1,t_2+2, \ldots, T$}
          \IF {$t=\tau_i$}
            \STATE Set $\tau_{i+1} = \minlength(2, \tau_i)$.
            \STATE Add $\Off(\tau_{i+1}-1)$ to $\sol$.
            \STATE Increment $i$.
      \ENDIF
\ENDFOR
\end{algorithmic} 
\end{algorithm} 
}
\begin{algorithm}
\caption{\pdbl}
\label{Alg: online_input_seq_with_thresh}
 {\bf Input:} The input sequence $\cI$ and prediction $\hat{T}$. \\
 {\bf Output:} The online solution $\sol$. \\
 
        Set $\sol:= \emptyset$, 
        $t_1:= \minlength(\epsilon/5, \hat{T})$, and 
        $t_2:= \maxlength(1+\epsilon/5, \hat{T})$. \\
        {\bf Phase 1:} Execute \dbl while $t < t_1$. \\
        {\bf Phase 2:} At $t = t_1$, add $\Off(t_2)$ to $\sol$. \\
        {\bf Phase 3:} If $t > t_2$, resume \dbl as follows: \\
        Set $i:=0$, $\tau_0:= t_2+1$. \\
        \textbf{for} $t=t_2+1,t_2+2, \ldots, T$ \\
        \hspace*{10pt} \textbf{if} $t=\tau_i$ \\
        \hspace*{10pt} \hspace*{10pt} Set $\tau_{i+1} = \minlength(2, \tau_i)$. \\
        \hspace*{10pt} \hspace*{10pt} Add $\Off(\tau_{i+1}-1)$ to $\sol$. \\
        \hspace*{10pt} \hspace*{10pt} Increment $i$.
\end{algorithm} 

As described in the introduction, the desiderata for an online algorithm with predictions are its consistency and robustness; we establish the tradeoff between these parameters for the \pdbl algorithm below.
\begin{theorem}\label{thm:robust-consistent}
The \pdbl algorithm has a consistency of $1+\epsilon$ and robustness of $5\left(1+\frac{1}{\epsilon}\right)$.
\end{theorem}
\begin{proof}
When the prediction is correct, i.e., $T = \hat{T}$, the algorithm only runs Phases 1 and 2. At the end of Phase 1, by Theorem~\ref{thm:dbl}, the cost of $\sol$ is at most $4\cdot \opt(t_1) \le (4\epsilon/5)\cdot \opt(\hat{T})$. In Phase 2, the algorithm buys a single solution of cost at most $(1 + \epsilon/5)\cdot \opt(\hat{T})$. Adding the two, and noting that the optimal cost is $\opt(\hat{T})$, we get a consistency bound of $1+\epsilon$.

For robustness, we consider three cases. First, if $t < t_1$, then the competitive ratio is $4$ by Theorem~\ref{thm:dbl}. Next, if $t > t_2$, then the total cost of $\sol$ is at most $(1+\epsilon) \opt(T)$ in Phases 1 and 2 (from the consistency analysis above), and at most $4\cdot \opt(T)$ in Phase 3 by Theorem~\ref{thm:dbl}. Thus, in this case, the competitive ratio is $5+\epsilon$. Finally, we consider the case $t_1 \le t \le t_2$. Here, the algorithm runs Phases 1 and 2, and the cost of $\sol$ is at most $(1+\epsilon)\cdot \opt(\hat{T})$ by the consistency analysis above. By monotonicity, the optimal solution is smallest when $T = t_1$, i.e., $\opt(T) \ge \frac{\epsilon}{5}\cdot \opt(\hat{T})$. Thus, the competitive ratio is bounded by $5\left(1 + \frac{1}{\epsilon}\right)$.
\end{proof}

We also show that this tradeoff between $(1+\epsilon)$-consistency and $O(1/\epsilon)$-robustness bounds is essentially tight.

\begin{theorem}\label{thm:cons_robus_trade}
Any algorithm for the \OS problem with predictions that has a consistency bound of $1+\epsilon$ must have a robustness bound of $\Omega\left(\frac{1}{\epsilon}\right)$. 
\end{theorem}
\begin{proof}
If $T\ge \hat{T}$, the algorithm has to buy a solution that is feasible for $\hat{T}$ at some time $\tau \le \hat{T}$. In particular, we must have $\opt(\tau) \le \epsilon \cdot \opt(\hat{T})$ for deterministic algorithms, else the consistency bound would be $> 1+\epsilon$ simply based on being feasible for $t = \tau$ which incurs cost $> \epsilon \cdot \opt(\hat{T})$ and again for $t = \hat{T}$ which incurs an additional cost of $\opt(\hat{T})$. This implies a robustness bound of $\Omega\left(\frac{1}{\epsilon}\right)$ if the input $T = \tau$. The same argument extends to randomized algorithms: now, since $\ex[\opt(\tau)] \le \epsilon \cdot \opt(\hat{T})$, it follows that $\ex\left[\frac{\opt(\hat{T})}{\opt(\tau)}\right] \ge \frac{\opt(\hat{T})}{\ex[\opt(\tau)]} = \Omega\left(\frac{1}{\epsilon}\right)$.

\end{proof}

Having shown the consistency and robustness of the \pdbl algorithm, we now analyze how its competitive ratio varies with error in the prediction $\hat{T}$. In particular, the next lemma shows that the competitive ratio gracefully degrades with prediction error for small error, and is capped at $4$ for large error.
\begin{lemma}\label{lemma: graceful_degradation}
Given a prediction $\hat{T}$ for the input length, the competitive ratio of \pdbl is given by:
\begin{equation*}   
    \comp 
    \le \begin{cases} 4 \text{ , } T \le t_1\\
    (1+\epsilon)\cdot \frac{\Off(\hat{T})}{\Off(T)} \text{ , } t_1 \le T \le t_2\\
    4 \text{ , } T > t_2
    \end{cases}
\end{equation*}
where $t_1$ represents the minimum value of $t$ that satisfies $\Off(t) \ge \frac{\epsilon}{5} \cdot \Off(\hat{T})$ and $t_2$ represents the maximum value of $t$ that satisfies $\Off(t) \le (1+\frac{\epsilon}{5}) \cdot\Off(\hat{T})$.
\end{lemma}
\begin{proof}
When $T\leq t_1$, the competitive ratio of $4$ follows from the doubling strategy of the algorithm.
Next, when $t_1 < T \le t_2$, the algorithm pays at most 
$4\cdot \frac{\epsilon}{5}\cdot \Off (\hat{T})$ until $t = t_1$ and then pays at most $\left(1+\frac{\epsilon}{5}\right)\cdot \Off(\hat{T})$ for the solution $\opt(t_2)$, which adds up to at most $(1+\epsilon)\cdot \Off (\hat{T})$. In contrast, the optimal cost is $\Off(T)$; hence, the competitive ratio is $(1+\epsilon)\cdot \frac{\Off(\hat{T})}{\Off(T)}$.
Finally, when $T > t_2$, then let $\tau_j \le T < \tau_{j+1}$ (using the notation in Algorithm~\ref{Alg: online_input_seq_with_thresh}).
The algorithm pays at most $$\left(1+\epsilon + 2 + \ldots 2^{j+1}\right)\Off(\hat{T}) 
\le 2^{j+2}\cdot \Off(\hat{T}),$$ 
while the optimal cost is at least $2^{j}\cdot \Off(\hat{T})$. Hence, the competitive ratio is at most $ 4$. 
\end{proof}

\section{{\sc Learn to Search}: A Regression Approach}
\label{sec:whitebox}
In the previous section, we designed an algorithm for the \OS problem that utilizes ML predictions. Now we delve deeper into how we can generate these predictions. More generally, we develop a regression-based approach to learn to solve an \OS problem. For this purpose, we first introduce some standard terminology for our learning framework, which we call the \ltos problem.

\subsection{Preliminaries}

An  instance $(x,z)$ of the \ltos  problem is given by a feature $x \in \mathbb{X}$, and the (unknown) cost of the optimal offline solution $z \in [1,M]$. The two quantities $x$ and $z$ are assumed to be drawn from a joint distribution on $\mathbb{X} \times [1, M]$. A prediction strategy works with a hypothesis class $\mathcal{F}$ that is a subset of functions $\mathbb{X}\mapsto [1, M]$ and tries to obtain the best function $f \in \mathcal{F}$ that predicts the target variable $z$ accurately. For notational convenience, we set our target $y = \ln z$, i.e., we try to predict the log-cost of the optimal solution. Note that predicting the log-cost of $\Off(T)$ is equivalent to predicting the input length $T$.\footnote{When multiple input lengths might have the same optimal cost, we can just pick the longest one.} Furthermore, let $\mathbb{D}$ denote the input distribution on $\mathbb{X}\times \mathbb{Y}$, where $\mathbb{Y} = [0, H]$ and $H = \ln M$; i.e., we assume that $(x,y) \sim \mathbb{D}$. 

We define a \ltos algorithm $\mathcal{A}$ as a strategy that receives a set of $m$ samples $S \sim \mathbb{D}^{m}$ for {\em training}, and later, when given the feature set $x$ of a {\em test} instance $(x, y) \sim \mathbb{D}$ (where $y$ is not revealed to the algorithm), it defines an online algorithm for the \OS problem with input $y$. Recall that an online algorithm constitutes a sequence of solutions that the algorithm buys at different times of the input sequence (see Algorithm ~\ref{Alg: gen_LTS} for a generic description of an \ltos algorithm).

\begin{algorithm}[ht]
\caption{A general \ltos algorithm} 
\label{Alg: gen_LTS}
\textbf{Training:}
 Given a Sample Set $S$, the training phase outputs a mapping $M$ from every feature vector $x\in \mathbb{X}$ to an increasing sequence of positive integers\\
 \textbf{Testing:}
 Given unknown sample $x\in \mathbb{X}$, define thresholds $M(x) = (\tau_0, \tau_1 \ldots )$\\
 Set $i := 0, \sol:= \Off(\tau_0-1)$.\\ 
 \textbf{while} (Input has not ended) \\ 
 {\hspace*{10pt}} {\bf if} (\sol is infeasible) \\
 \hspace*{20pt}  $\sol := \Off(\tau_{i+1}-1)$. \\
 \hspace*{20pt} Increment $i$. 
\end{algorithm}


We will use the notation $\comp_{\mathcal{A}}(x,y)$ to denote the competitive ratio obtained by an algorithm $\mathcal{A}$ on the instance $(x,y)$.
For a given set of thresholds $(\tau_0, \tau_1 \ldots )$, define $i_{T} = \min_{\tau_i > T} i$. Then, $\mathcal{A}$ pays a total cost of $\sum_{i=0}^{i_{T}}\Off (\tau_i)$, and thus the competitive ratio is $$\comp_{\mathcal{A}}(x,y)=\frac{\sum_{i=0}^{i_T}\Off (\tau_i)}{e^{y}}.$$ 
We define the ``efficiency'' of a \ltos algorithm by comparing its performance with the best achievable competitive ratio. The optimal competitive ratio for a given distribution may be strictly greater than $1$. For example, consider the distribution where $x$ is fixed (say $x_0$) and $z$ is uniformly distributed over the set $\{2,4\}$. One can verify that the best strategy for the above distribution is to buy the solution of cost $2$, and then if the input has not ended, then buy the solution of cost $4$. The competitive ratio for this strategy (in expectation) is $1.25$.
\begin{definition}
A \ltos algorithm $\mathcal{A}$ is said to be $\epsilon$-efficient if $$\ex_{(x,y)\sim \mathbb{D}}\comp_{\mathcal{A}}(x,y) \leq \rho^{*} + \epsilon,$$ 
where $\rho^{*} = \ex_{(x,y)\sim \mathbb{D}}\comp_{\mathcal{A}^{*}}(x,y)$ and $\mathcal{A}^{*}$ is an optimal solution that has full knowledge of $\mathbb{D}$ and no computational limitations.
\end{definition}

The ``expressiveness'' of a function family is  captured by the following standard definition:
\begin{definition}
A set $S = \{x_1, x_2, \ldots x_m\}$ is said to be ``shattered'' by a class $\mathcal{F}$ of real-valued functions $S \mapsto [0,H]$ if there exists ``witnesses'' $R = \{r_1, r_2 \ldots r_m \} \in [0,H]^{m}$ such that the following condition holds:
For all subsets $T \subseteq S$, there exists an $f \in \mathcal{F}$ such that $f(x_i) > r_i$ if and only if $x_i \in T$. The ``pseudo-dimension'' of $\mathcal{F}$ (denoted as $\pdim(\mathcal{F})$) is the cardinality of the largest subset $S\subseteq X$ that is shattered by $\mathcal{F}$.
\end{definition}

\subsection{The Sample Complexity of {\sc LearnToSearch}}

\begin{figure}[ht]
    \centering
    \includegraphics[width=0.45\textwidth]{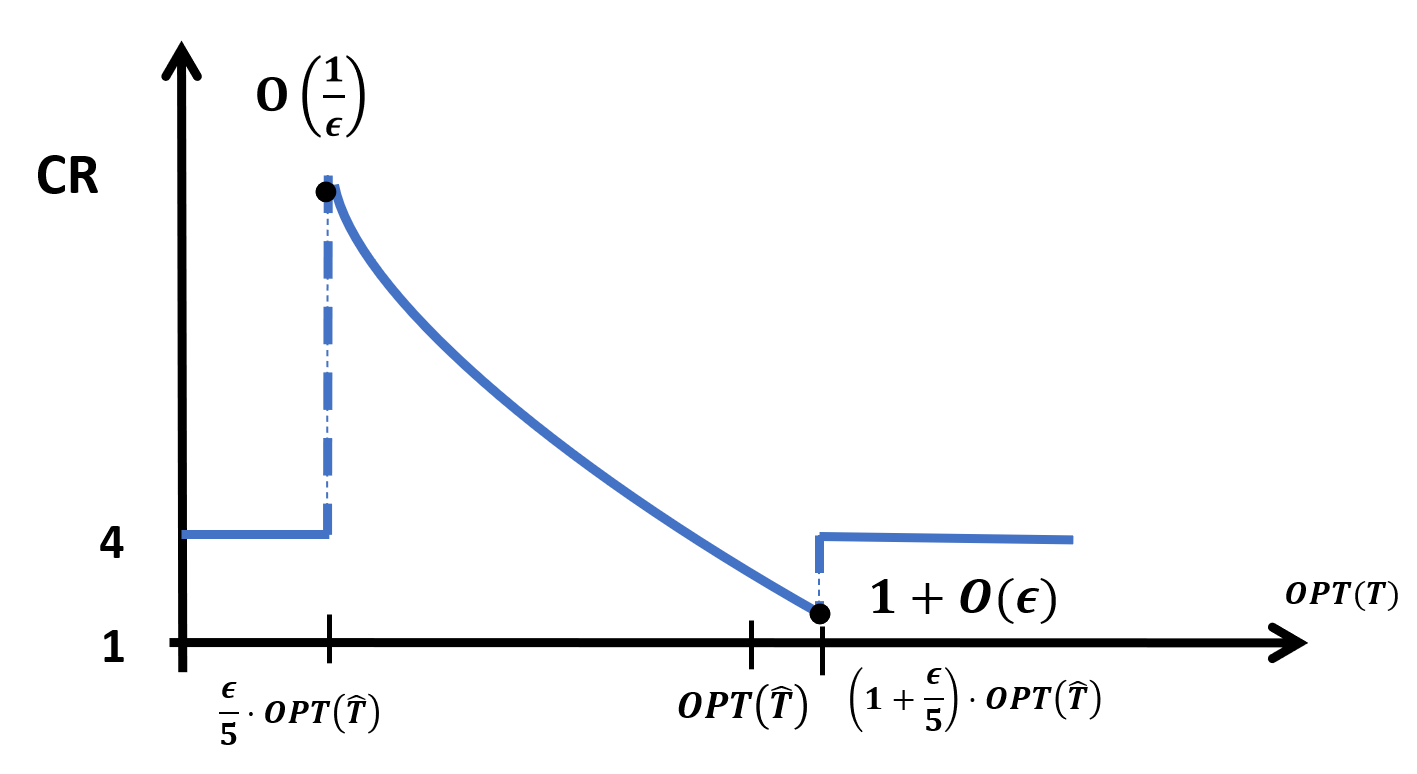}
    \caption{Competitive ratio of the \pdbl algorithm for a fixed prediction $\hat{T}$ as a function of the input $T$, where the prediction is $\hat{T}$}
    \label{fig:cr}
\end{figure}

Our overall strategy is to learn a suitable predictor function $f\in \mathcal{F}$ and use $f(x)$ as a prediction in the \pdbl algorithm. Note that prediction errors on the two sides (over- and under-estimation) affect the competitive ratio of \pdbl  (given by Lemma~\ref{lemma: graceful_degradation}) in different ways. If we underestimate $\opt(T)$ by a factor less than $1+\frac{\epsilon}{5}$, i.e., $\opt(T') \leq \opt(T) \leq \left( 1 + \frac{\epsilon}{5} \right) \cdot \opt(T')$, the competitive ratio remains $1+O(\epsilon)$, but a larger underestimate causes the competitive ratio to climb up to 4. On the other hand, if we overestimate $\opt(T)$, then the competitive ratio grows steadily by the ratio of over-estimation, until it reaches $5\cdot\left(1+\frac{1}{\epsilon}\right)$, beyond which it drops down (and stays at) 4. This asymmetric dependence 
is illustrated in Figure~\ref{fig:cr}.

At a high level, our goal is to use regression to obtain the best function $f\in \mathcal{F}$. But, the asymmetric behavior of the competitive ratio suggests that we should not use a standard loss function in the regression analysis. 
Let $\epsilon$ be the accuracy parameter for the \pdbl algorithm, and let $\hat{y} = \ln \opt(\hat{T})$ and $y = \ln \opt(T)$ be the predicted and actual log-cost of the optimal solution respectively. Then we define the following loss function that follows the asymmetric behaviour of the competitive ratio for \pdbl:

\begin{definition}\label{def: comp_loss}
The $\epsilon$-parameterized competitive error is defined as:
$$
\ell_{\epsilon}(y,\hat{y}) = \begin{cases} 
\frac{5}{\epsilon} - 1 \text{ when $y \leq \hat{y} - \ln \frac{5}{\epsilon}$}\\
e^{\hat{y} -  y} -1 \text{ when $\hat{y} - \ln \frac{5}{\epsilon} < y \le \hat{y}$}\\
\frac{1}{\epsilon}\cdot (y - \hat{y}) \text{ when $\hat{y} < y \le \hat{y} + \ln \left(1 + \frac{\epsilon}{5}\right)$}\\
1 \text{ when $y > \hat{y}+\ln\left(1+\frac{\epsilon}{5}\right)$}.
\end{cases}$$
\end{definition}

We give more justification for using this loss function, and show that standard loss functions do not suffice for our purposes in Section~\ref{sec:traditional}. Using this loss function, we can measure the error of a function for an input distribution or for a fixed input set:

\begin{definition}
Given a distribution $\mathbb{D}$ on the set $\mathbb{X}\times \mathbb{Y}$ and function $f: \mathbb{X} \mapsto \mathbb{Y}$, we define
$$\err_{\mathbb{D}, \epsilon}(f) = \ex_{(x,y)\sim \mathbb{D}}[\ell_\epsilon (y,f(x))].$$
Alternatively, for a set of samples, $S \sim \mathbb{D}^{m} $, we define,
$$\err_{S, \epsilon}(f) = \frac{1}{m}\cdot\sum_{i=1}^{m}\ell_\epsilon (y_i,f(x_i)).$$
\end{definition}

Our high-level goal is to use samples to optimize for the loss function called $\epsilon$-parameterized competitive error that we defined above over the function class $\mathcal{F}$, and then use an algorithm that translates the empirical error bound to a competitive ratio bound. This requires, in the training phase, that we optimize the empirical loss on the training samples. We define such a minimizer below:


\begin{definition}
For a given set of samples $S \sim \mathbb{D}$ and a function family $\mathcal{F}$, we denote an optimization scheme $\mathcal{O}: S \mapsto \mathcal{F}$ as $\epsilon-$Sample Error Minimizing (SEM) if it returns a function $\hat{f} \in \mathcal{F}$ satisfying:
$$\err_{S, \epsilon}(\hat{f}) \le \inf_{f\in \mathcal{F}}\left[\err_{S, \epsilon}(f)\right] + \epsilon.$$
\end{definition}
For the rest of this paper, we will assume that we are given an $\epsilon-$ SEM routine for arbitrary $\epsilon > 0$.
We are now ready to present our \ltos algorithm (Algorithm~\ref{Alg: online_cost_with_pred_params}), which basically uses the predictor with minimum expected loss to make predictions for \pdbl.

\begin{algorithm}
\caption{A \ltos algorithm with accuracy parameter $\epsilon$}
\label{Alg: online_cost_with_pred_params}
 \textbf{Training:}\\
 \hspace*{10pt}{\bf Input:} Sample Set $S$, Function Family $\mathcal{F}$\\
 \hspace*{10pt}{\bf Output:} $\hat{f}$ output by an $\epsilon$-SEM algorithm $\mathcal{O}$, i.e.,
 $\err_{S, \epsilon}(\hat{f}) \le \inf_{\tilde{f} \in \mathcal{F}} \err_{S, \epsilon}(\tilde{f}) + \epsilon$.\\
 \textbf{Testing:}\\
 \hspace*{10pt} Given new sample $x$, set $\hat{y} = \hat{f}(x)$.\\
 \hspace*{10pt} Predicted prefix length: $\hat{T} = \max_{\Off(t)\le e^{\hat{y}}} t$.\\
 \hspace*{10pt} Call \pdbl with $\hat{T}$ and $\epsilon$.
\end{algorithm}
\vspace{-5pt}
We relate the competitive ratio of Algorithm~\ref{Alg: online_cost_with_pred_params} to the error of function $\hat{f}$ obtained during training:
\begin{lemma}\label{lemma: learner_to_alg}
Algorithm~\ref{Alg: online_cost_with_pred_params} has a competitive ratio upper bounded by $\Big(1+ \epsilon + 3\cdot \err_{\mathbb{D},\epsilon}(\hat{f})\Big)$.
\end{lemma}
\begin{proof}
We use Lemma~\ref{lemma: graceful_degradation} to prove this result. Let $\hat{y}$ and $\hat{T}$ be as in the description of Algorithm~\ref{Alg: online_cost_with_pred_params}. Let $t_1, t_2$ be as in the statement of Lemma~\ref{lemma: graceful_degradation} with respect to $\hat T$. 

Now consider the following cases (as in the statement of Lemma~\ref{lemma: graceful_degradation}) 
\begin{itemize}
    \item $T < t_1$: by definition of $t_1$, 
    $\opt(T) < \frac{\epsilon}{5} \cdot \opt(\hat{T})$, and so, $y < \hat{y} - \ln \frac{5}{\epsilon}.$ Since the competitive ratio is at most 4 in this case, we see  that (using Definition~\ref{def: comp_loss}) this is at most $\ell_{\epsilon}(\hat{y},y).$ 
    \item $t_1 \leq T \leq \hat{T}$: In this case, $\hat{y} - \ln \frac{5}{\epsilon} \leq y \leq \hat{y}$.  This case, the competitive ratio is at most 
    $$ (1 + \epsilon) \cdot e^{\hat{y} - y} \leq  1 + \epsilon + 3 \left( e^{\hat{y}-y}-1 \right) = 1 + \epsilon + 3\ell_{\epsilon}(\hat{y},y), $$
    where the first inequality follows from the fact that $\hat{y} \geq y$. 
    \item $\hat{T}  \leq T \leq t_2:$  Here $\hat{y} < y \leq \hat{y} + \ln \left( 1 + \frac{\epsilon}{5} \right). $
    Again, the competitive ratio is at most 
    $$ (1 + \epsilon) \cdot e^{\hat{y} - y} \leq  1 + \epsilon \le 1 + \epsilon + 3\ell_{\epsilon}(\hat{y},y),   $$
    where the first inequality follows from $\hat{y} < y$. 
    \item $T > t_2$: Here $ y \geq \hat{y} + \ln \left( 1 + \frac{\epsilon}{5} \right)$. Lemma~\ref{lemma: graceful_degradation} shows that the competitive ratio is at most 4, which is at most $1 + \epsilon + 3 \ell_{\epsilon}(\hat{y},y).$
\end{itemize}

We note that for all values $y$, the competitive ratio is upper bounded by $1+\epsilon + 3\cdot\ell_{\epsilon}(\hat{y},y)$, where $\ell_{\epsilon}(\hat{y},y)$ is the $\epsilon$-parameterized competitive error of $\hat{y}$. So, the expected competitive ratio is $\le 1+\epsilon + 3\cdot \err_{\mathbb{D},\epsilon}(\hat{f})$.
\end{proof}

\paragraph{Standard and Agnostic Models.}

We consider two different settings. First, we assume that the function class $\mathcal{F}$ contains the function $f^*$ that maps the feature set $x$ to $y$ -- we call this the {\em standard} model. We relax this assumption in the more general {\em agnostic} model, where the function class $\mathcal{F}$ is arbitrary. 
%
%
%
%
%
%
In terms of the error function, in the standard model, we have $\inf_{f\in \mathcal{F}} \err_{\mathbb{D}, \epsilon}(f) = \inf_{f\in \mathcal{F}} \err_{S,\epsilon}(f) = 0$, while no such guarantee holds in the agnostic model.

\subsection{Analysis in the Standard Model}

Next, we analyze the competitive ratio of Algorithm~\ref{Alg: online_cost_with_pred_params} in the standard model, i.e., when $\inf_{f\in \mathcal{F}} \err_{\mathbb{D}, \epsilon}(f) = \inf_{f\in \mathcal{F}} \err_{S, \epsilon}(f) = 0$. 



\begin{theorem}\label{thm:restricted}
In the standard model, Algorithm~\ref{Alg: online_cost_with_pred_params} obtains a competitive ratio of $1+O(\epsilon)$ with probability at least $1-\delta$, when using  $O\left(\frac{H\cdot d \log \frac{1}{\epsilon} \log \frac{1}{\delta}}{\epsilon}\right)$ samples, where $d=\pdim(\mathcal{F})$. 
\end{theorem}

When the cost of the optimal solution $\Off(\tau)$ is hard to compute, we can replace the offline optimal with an online algorithm that achieves competitive ratio $c$ given the value of $\tau$ to get the following:

\begin{corollary}\label{thm:compeitive}
In the standard model, if there exists a $c$-competitive algorithm for $\Off(\tau)$ given the value of prefix-length $\tau$, Algorithm~\ref{Alg: online_cost_with_pred_params} obtains a competitive ratio of $c(1+O(\epsilon))$ with probability at least $1-\delta$, when using  $O\left(\frac{H\cdot d \log \frac{1}{\epsilon} \log \frac{1}{\delta}}{\epsilon}\right)$ samples, where $d=\pdim(\mathcal{F})$. 
\end{corollary}

We also show that the result in Theorem~\ref{thm:restricted} is tight up to a factor of $H\log 1/\epsilon$:

\begin{theorem}
\label{thm:restlower}
Let $\mathcal{F}$ be a family of real valued functions such that there exists a function $f^{*}:\mathbb{X}\mapsto \mathbb{Y}$ that $f^{*}(x)=y$ and let $d=\pdim(\mathcal{F})$. There exists an instance of the \ltos problem that enforces any algorithm to query $\Omega\left(\frac{d \log \frac{1}{\delta}}{\epsilon}\right)$ samples in order to have an expected competitive ratio of $1+\epsilon$ with probability $\ge 1-\delta$.
\end{theorem}
In order to have sample complexity bounds relating to the pseudo dimension of the function class, we would need to introduce the notion of covering numbers and relate them to the pseudo-dimension.

\begin{definition}
Given a set $S$ in Euclidean space and a metric $d(\cdot,\cdot)$, the set $W \subseteq S$ is said to be $\epsilon$ cover of $S$ if for any $s \in S$, there exists a $w \in W$ such that $d(s,w) \le \epsilon$. The smallest possible cardinality of such an $\epsilon$ cover is known as the $\epsilon$ covering number of $S$ with respect to $d$ and is denoted as $\mathcal{N}_{d(\cdot,\cdot)}(\epsilon, S)$.
\end{definition}

When $d$ is given by the distance metric 
$$d_{p}(r,s) = \big|\sum_{i=1}^{d}(r_i - s_i)^{p}\big|^{1/p},$$ where $r = (r_1, r_2 \ldots r_{d}), s= (s_1, s_2 \ldots s_{d}) \in \mathbb{R}^{d}$, we shall denote the $\epsilon$ covering number of a set $S$ by $\mathcal{N}_{p}(\epsilon, S)$. For a given real-valued function family $\mathcal{F}$ and $x = \left(x_1,x_2, \ldots, x_{m} \right) \in \mathbb{X}^{m}$, we denote $$\mathcal{F}_{\mid x} = \{\left(f(x_1), f(x_2), \ldots, f(x_m) \right)\mid f\in \mathcal{F} \} \quad \text{and}$$ $$\mathcal{N}_{p}(\epsilon, \mathcal{F}, m) = \sup_{x \in \mathbb{X}^{m}}\left[\mathcal{N}_{p}(\epsilon, \mathcal{F}_{\mid x}) \right].$$ Note that
$\mathcal{N}_{1}(\epsilon, \mathcal{F}, m) \le \mathcal{N}_{2}(\epsilon, \mathcal{F}, m) \le \mathcal{N}_{\infty}(\epsilon, \mathcal{F}, m)$.

Given a loss function $\ell(\cdot, \cdot)$, and sample set $S = \{(x_i,y_i), i={1,2\ldots m}\}$, we can define $(\ell_{\mathcal{F}})_{\mid S}$ as :
\begin{equation*}
    (\ell_{\mathcal{F}})_{\mid S} = \{\ell_{f}(x_i, y_i) | (x_i,y_i) \in S, f \in \mathcal{F}\} \subset \mathbb{R}^{m}.  
\end{equation*}

where $\ell_{f}(x_i, y_i) = \ell(y_i, f(x_i))$.

The following is a well-known result that  relates covering numbers to the pseudo dimension (cf. Theorem 12.2 in Book~\cite{anthony1999learning}):
\begin{lemma}\label{lemma: bound_covering_number}
Let $\mathcal{F}$ be a real-valued function family with pseudo dimension $d$, then for any $\epsilon \le \frac{1}{d}$, we have
$$\mathcal{N}_{1}\left( \epsilon, \mathcal{F}, m  \right) \le O\left(\frac{1}{\epsilon^d}\right).$$
\end{lemma}

First, we relate covering numbers to the difference  between
$\err_{S, \epsilon}(f)$ and $\err_{\mathbb{D}, \epsilon}(f)$. This will be crucial in proving Theorem~\ref{thm:restricted}.
\begin{lemma}\label{lemma: bound_bad_prob}
Let $\mathbb{D}$ be a distribution on $\mathbb{X}\times \mathbb{Y}$ and let $S \in \mathbb{D}^{m}$. For $0 \le \eta \leq 12$ and $m\ge \frac{8\cdot H}{\eta^2}$, for any real valued function family $\mathcal{F}$,  we have:
\begin{align*}
\mathbb{P}_{S \in \mathbb{D}^{m}}\left[\err_{\mathbb{D}, \epsilon}(f) \le (1+\alpha)\cdot\err_{S, \epsilon}(f) + \eta^2\cdot\left(1+\frac{1}{\alpha}\right) \right] 
\le 4\cdot\mathcal{N}_1\left(\frac{\eta\epsilon}{8}, \mathcal{F}, 2m\right) \cdot \exp\left(-\frac{m\cdot \eta^2 \cdot }{64H}\right).
\end{align*}

Moreover, we also have the other side as :
\begin{equation*}
\mathbb{P}_{S \in \mathbb{D}^{m}}\left[\err_{S, \epsilon}(f) \le \frac{2\alpha+1}{(1+\alpha)}\cdot\err_{\mathbb{D}, \epsilon}(f) + \eta^2\cdot\left(1+\frac{1}{\alpha}\right) \right] 
\le 4\cdot\mathcal{N}_1\left(\frac{\eta\epsilon}{8}, \mathcal{F}, 2m\right) \cdot \exp\left(-\frac{m\cdot \eta^2 \cdot }{64H}\right)
\end{equation*}

\end{lemma}










To prove this lemma, we need the following definition.

\begin{definition}
The normalised ($\epsilon$ parameterised) error is defined as :
\begin{equation*}
    \hat{\err}_{S,\mathbb{D}, \epsilon}(f) = \frac{\abs{\err_{S, \epsilon}(f) - \err_{\mathbb{D}, \epsilon}(f)}}{\sqrt{\err_{\mathbb{D}, \epsilon}(f)}}.
\end{equation*}
\end{definition}


\begin{lemma}\label{lemma: bound_bad_prob2}
Let $\mathbb{D}$ be a distribution on $\mathbb{X} \times \mathbb{Y}$ and let $S\sim \mathbb{D}^{m}$. For $\eta \le 12$, and $m\ge \frac{8H}{\eta^2}$, for any real valued function family $\mathcal{F}$, we have
\begin{equation*}
\mathbb{P}_{S\sim \mathbb{D}^{m}}\left[\sup_{f\in \mathcal{F}}\abs{\hat{\err}_{S}(f) - \hat{\err}_{\mathbb{D}}(f)} \ge \eta\right] \le 4\cdot \mathcal{N}\left(\eta/8, \ell_{\mathcal{F}}, 2m\right) \cdot exp\left(-\frac{m\cdot \eta^2}{64H}\right).
\end{equation*}
\end{lemma}

We break this proof into four separate claims as illustrated below.

First, we reduce the probability of the event:
$\left[\hat{\err}_{S}(f) \ge \eta\right]$ to a probability term involving two sample sets $S, \bar{S}$ the members of which are drawn independently.

\begin{lemma}\label{lemma: symmetrization}
For $m \ge \frac{8H}{\eta^2}$, we have that:
$$\mathbb{P}_{S \sim \mathbb{D}^{m}}\left[\sup_{f\in \mathcal{F}} \hat{\err}_{S}(f)\ge \eta\right] 
\le 2\cdot  \mathbb{P}_{(S,\bar{S})\sim D^{m} \times D^{m}}\left[\sup_{f\in \mathcal{F}} \abs{\hat{\err}_{S}(f) - \hat{\err}_{\bar{S}}(f)} \ge \eta/2\right].$$
\end{lemma}


\begin{proof}

For a given sample $S\sim \mathbb{D}^{m}$, let $f^{S}_{bad} \in \mathcal{F}$ denote a function $f$ such that $\abs{\hat{\err}_{S}(f) - \hat{\err}_{\mathbb{D}}(f)} \ge \eta$ if it exists, otherwise we set $f$ to be any fixed function in the family $\mathcal{F}.$ Now, 

\begin{align*}
    \mathbb{P}_{(S,\bar{S})\sim D^{m} \times D^{m}}  \left[\sup_{f\in \mathcal{F}}\abs{\hat{\err}_{S}(f) - \hat{\err}_{\bar{S}}(f)} \ge \frac{\eta}{2}\right]
    &\ge \mathbb{P}_{(S,\bar{S})\sim D^{m} \times D^{m}} \left[\abs{\hat{\err}_{S}(f^{S}_{bad}) - \hat{\err}_{\bar{S}}(f^{S}_{bad})} \ge \frac{\eta}{2}\right]\\
    &\ge \mathbb{P}_{(S,\bar{S})\sim D^{m} \times D^{m}} \left[\left\{\hat{\err}_{S}(f^{S}_{bad})\ge \eta \right\}\cap \left\{\hat{\err}_{\bar{S}}(f^{S}_{bad})\le \frac{\eta}{2}\right\}\right]\\
    &=\mathbb{P}_{S\sim \mathbb{D}^{m}} \left[\hat{\err}_{S}(f^{S}_{bad})\ge \eta\right] \cdot \mathbb{P}_{\bar{S}\sim \mathbb{D}^{m} \mid S}\left[\hat{\err}_{\bar{S}}(f^{S}_{bad})\le \frac{\eta}{2}\right]\\
    &=\mathbb{P}_{S\sim \mathbb{D}^{m}} \left[\sup_{f\in \mathcal{F}}\hat{\err}_{S}(f)\ge \eta\right] \cdot \mathbb{P}_{\bar{S}\sim \mathbb{D}^{m} \mid S}\left[\hat{\err}_{\bar{S}}(f^{S}_{bad})\le \frac{\eta}{2}\right].
\end{align*}

Now, the term  $\mathbb{P}_{\bar{S}\sim \mathbb{D}^{m} \mid S}\left[\hat{\err}_{\bar{S}}(f^{S}_{bad})\le \frac{\eta}{2}\right]$ is bounded below by the Chebyshev's inequality as follows:
\begin{equation*}
\mathbb{P}_{\bar{S}\sim \mathbb{D}^{m} \mid S}\left[\hat{\err}_{\bar{S}}(f^{S}_{bad})\le \frac{\eta}{2}\right]
\ge 1 - \frac{{\rm Var}_{\bar{S} \sim \mathbb{D}^{m}\mid S}\left[\hat{\err}_{\bar{S}}(f^{S}_{bad}) \right]}{\frac{\eta^2}{4}}
\ge 1 - \left(\frac{H}{m \cdot \frac{\eta^2}{4} }\right)
\ge \frac{1}{2},
\end{equation*}
where the last inequality follows from the fact that $m \ge \left( \frac{8\cdot H}{\eta^2} \right)$ and that
\begin{align*}
    \frac{Var(\err_{\bar{S}}(f^{S}_{bad}))}{\err_{\mathbb{D}}(f^{S}_{bad})} = \frac{1}{m} \cdot \sum_{i=1}^{i=m} \frac{Var(\ell_{f^{S}_{bad}}(x_i, y_i)))}{\mathbb{E}(\ell_{f^{S}_{bad}}(x_i, y_i))}
    \le \frac{1}{m}\cdot H,
\end{align*}
because  $\ell_{f^{S}_{bad}}(x_i, y_i)) \in [0, H]$. 

%
\end{proof}
The second claim intuitively says that the probabilities remain unchanged under symmetric permutations. Let $\sigma$ denote a permutation on the set $\{ 1, 2, \ldots, 2m\}$
such that for each $i\in \{1, 2, \ldots, m\}$, we use either of the two mappings:
\begin{itemize}
    \item $\sigma(i) = i$ and $\sigma(m+i) = m+i$, or
    \item $\sigma(i) = m+i$ and $\sigma(m+i) = i$.
\end{itemize}
Let $\Gamma ^m$ denote the set of all such permutations $\sigma$.  Suppose we draw i.i.d. samples $S\sim\mathbb{D}^m$ and $\bar{S}\sim\mathbb{D}^m$; let $S = \{s_1, s_2, \ldots, s_m\}$ and $\bar{S} = \{\bar{s}_1, \bar{s}_2, \ldots, \bar{s}_m\}$. Then, define $\sigma(S)$ and $\sigma(\bar{S})$ by using a permutation $\sigma\in\Gamma^m$ as follows. Let $\sigma(S) = \{s'_1, s'_2, \ldots, s'_m\}$ and $\sigma(\bar{S}) = \{\bar{s}'_1, \bar{s}'_2, \ldots, \bar{s}'_m\}$ such that $s'_i = s_i$ and $\bar{s}'_i = \bar{s}_i$ if $\sigma(i) = i$ and $\sigma(m+i) = m+i$, while $s'_i = \bar{s}_i$ and $\bar{s}'_i = s_i$ if $\sigma(i) = m+i$ and $\sigma(m+i) = i$. Let $U^m$ denote the uniform distribution over $\Gamma^m$.

\begin{lemma}\label{lemma: permutation}
For every $f \in \mathcal{F}$:
$$\mathbb{P}_{(S,\bar{S})\sim D^{m} \times D^{m}}\left[\abs{\hat{\err}_S(f) - \hat{\err}_{\bar{S}}(f)} \ge \eta/2\right] \le \sup_{(S,\bar{S}) \in (\mathbb{X}\times \mathbb{Y})^{2m}}\left(\mathbb{P}_{\sigma \sim U^m}\left[ \abs{\hat{\err}_{\sigma(S)} (f) - \hat{\err}_{\sigma(\bar{S})}(f)}\ge \eta/2 \right]\right).$$
\end{lemma}

\begin{proof}
We have for every $f\in \mathcal{F}$:
\begin{align*}
    \mathbb{P}_{(S,\bar{S})\sim D^{m} \times D^{m}}&\left[\abs{\hat{\err}_S(f) - \hat{\err}_{\bar{S}}(f)} \ge \eta/2\right]\\
    &=\mathbb{P}_{(S,\bar{S})\sim D^{m} \times D^{m},~\sigma \sim U^m}\left[ \abs{\hat{\err}_{\sigma(S)} (f) - \hat{\err}_{\sigma(\bar{S})}(f)}\ge \eta/2 \right] \quad \text{(by the i.i.d. property)}\\
    &\le \sup_{(S,\bar{S}) \in (\mathbb{X}\times \mathbb{Y})^{2m}}\left(\mathbb{P}_{\sigma \sim U^m}\left[ \abs{\hat{\err}_{\sigma(S)} (f) - \hat{\err}_{\sigma(\bar{S})}(f)}\ge \eta/2 \right]\right),
\end{align*}
where in the last expression we chose the members of $S, \bar{S}$ adversarially instead of randomly.
\end{proof}

Third, we make use of covering numbers to quantify the above probability.
\begin{lemma}\label{lemma: quantization}
Fix a $(S, \bar{S}) \in (\mathbb{X}\times \mathbb{Y})^{2m}$. Consider the set $\mathcal{G} \in \mathcal{F}$ such that $\ell_{\mathcal{G}}(S,\bar{S})$ is an $\frac{\eta}{8}$-covering (wrt $d_1(\cdot, \cdot)$) of the set $\ell_{\mathcal{F}}(S,\bar{S}) = \{\ell_{f}(x_i, y_i)\mid (x_i,y_i) \in S\cup\bar{S}, f \in \mathcal{F} \}\subset [0,H]^{2m}$. Then :
$$\mathbb{P}_{S\sim \mathbb{D}^{m}} \left[\sup_{f\in \mathcal{F}}\hat{\err}_S(f) \ge \eta\right]\le \mathcal{N}\left(\frac{\eta}{8}, \ell_{\mathcal{F}}, 2m\right) \cdot\max_{g \in \mathcal{G}} \mathbb{P}_{S\sim \mathbb{D}^{m}} \left[\abs{\hat{\err}_{\sigma(S)} (g) - \hat{\err}_{\sigma(\bar{S})}(g)}\ge \frac{\eta}{4}\right].$$
\end{lemma}
\begin{proof}

Note that the cardinality of $\mathcal{G}$ is less than $\mathcal{N}_{1}\left(\eta/8, \ell_{\mathcal{F}}, 2m \right)$ and is a bounded number. We claim that whenever an $f \in \mathcal{F}$ satisfies,
$\abs{\hat{\err}_{\sigma(S)} (f) - \hat{\err}_{\sigma(\bar{S})}(f)}\ge \frac{\eta}{2}$, then there exists a $g \in \mathcal{G}$ such that,
$\abs{\hat{\err}_{\sigma(S)} (g) - \hat{\err}_{\sigma(\bar{S})}(g)}\ge \frac{\eta}{4}$.

Let $g$ satisfy that,
$\frac{1}{2m} [ \sum_{i=1}^{2m}\abs{\ell_{g}(x_i,y_i) - \ell_{g}(x_i,y_i)} ] \leq \frac{\eta}{8}$.

We are guaranteed that such a $g$ exists, since it is in the cover.
\begin{align*}
    \frac{\eta}{2} &\le \abs{\hat{\err}_{\sigma(S)}(f) - \hat{\err}_{\sigma(\bar{S})}(f)  }\\
    &=\abs{ (\hat{\err}_{\sigma(S)}(f) - \hat{\err}_{\sigma(S)}(g) ) - (\hat{\err}_{\sigma(\bar{S})}(f) - \hat{\err}_{\sigma(\bar{S})}(g) ) +  (\hat{\err}_{\sigma(S)}(g) - \hat{\err}_{\sigma(\bar{S})}(g)) }\\
    &=\abs{ (\hat{\err}_{\sigma(S)}(f) - \hat{\err}_{\sigma(S)}(g) )} + \abs{\hat{\err}_{\sigma(\bar{S})}(f) - \hat{\err}_{\sigma(\bar{S})}(g) } +  \abs{\hat{\err}_{\sigma(S)}(g) - \hat{\err}_{\sigma(\bar{S})}(g)}\\
    &= \abs{\hat{\err}_{\sigma(S)}(g) - \hat{\err}_{\sigma(\bar{S})}(g)} + \abs{\frac{1}{m}\sum_{i=1}^{m}\left( \ell_{f}(x_i,y_i) - \ell_{g}(x_i,y_i) \right)} + \abs{\frac{1}{m}\sum_{i=m+1}^{2m}\left( \ell_{f}(x_i,y_i) - \ell_{g}(x_i,y_i) \right)}\\
    &\le \abs{\hat{\err}_{\sigma(S)}(g) - \hat{\err}_{\sigma(\bar{S})}(g)} + \frac{1}{m}\sum_{i=1}^{2m}\abs{\ell_{f}(x_{\sigma(i), y_{\sigma(i)}}) - \ell_{g}(x_{\sigma(i), y_{\sigma(i)}})}\\
    &< \abs{\hat{\err}_{\sigma(S)}(g) - \hat{\err}_{\sigma(\bar{S})}(g)} + \frac{\eta}{4}.
\end{align*}

Therefore, we get:
\begin{align*}
\mathbb{P}_{S\sim \mathbb{D}^{m}}& \left[\abs{\hat{\err}_{\sigma(S)} (f) - \hat{\err}_{\sigma(\bar{S})}(f)}\ge \eta/2 \right] \\
&\le \mathbb{P}_{S\sim \mathbb{D}^{m}} \left[\abs{\hat{\err}_{\sigma(S)} (g) - \hat{\err}_{\sigma(\bar{S})}(g)}\ge \eta/4\right]\\
&\le \abs{\mathcal{G}}\cdot \max_{g \in \mathcal{G}} \mathbb{P}_{S\sim \mathbb{D}^{m}} \left[\abs{\hat{\err}_{\sigma(S)} (g) - \hat{\err}_{\sigma(\bar{S})}(g)}\ge \eta/4\right].\\
&\le \mathcal{N}\left(\frac{\eta}{8}, \ell_{\mathcal{F}}, 2m\right) \cdot\max_{g \in \mathcal{G}} \mathbb{P}_{S\sim \mathbb{D}^{m}} \left[\abs{\hat{\err}_{\sigma(S)} (g) - \hat{\err}_{\sigma(\bar{S})}(g)}\ge \eta/4\right].\qedhere
\end{align*}
\end{proof}

Our final step is to bound $\mathbb{P}_{\sigma \sim U^{m} }\left[\abs{\hat{\err}_{\sigma(S)} (g) - \hat{\err}_{\sigma(\bar{S})}(g)}\ge \frac{\eta}{4} \right]$ for all $(S,\bar{S}) \in (\mathbb{X}\times \mathbb{Y})^{2m}$, which is effected by the last claim.

\begin{lemma}\label{lemma: bernstein}
For any $f\in \mathcal{F}$, with $\eta \le 12$:
$$\mathbb{P}_{\sigma \sim U^{m} }\left[\abs{\hat{\err}_{\sigma(S)} (f) - \hat{\err}_{\sigma(\bar{S})}(f)}\ge \frac{\eta}{4} \right] \le 2\cdot \exp\left(-\frac{m \cdot \eta^2}{64H}\right).$$
\end{lemma}
\begin{proof}
We use Bernstein's inequality \cite{craig1933tchebychef} that says for $n$ independent zero-mean random variables $X_i$'s satisfying $\abs{X_i}\le M$, we have:
$$\mathbb{P}\left(\abs{\sum_{i}^{n} X_i } > t \right) \le 2\cdot \exp\left({\frac{-\frac{t^2}{2}}{\sum_{i=1}^{n}\mathbb{E}[X^2_i]+\frac{1}{3}M\cdot t}}\right).$$

Note that the quantity  $\hat{\err}_{\sigma(S)} (f) - \hat{\err}_{\sigma(\bar{S})}(f)$ is simply an average of $m$ random variables, each of which has a variance upper bounded by $H$. Then applying the above bound:

\begin{align*}
    \mathbb{P}_{\sigma \sim U^{m}}
    &\left[\abs{\hat{\err}_{\sigma(S)} (f) - \hat{\err}_{\sigma(\bar{S})}(f)}\ge \frac{\eta}{4} \right]\\ &\le 2\cdot \exp\left(-\frac{m\cdot \eta^2 }{32H(1+\frac{\eta}{12})}\right)\\
    &\le 2\cdot \exp\left(-\frac{m \cdot \eta^2}{64H}\right). \qedhere
\end{align*}
\end{proof}
We will use the Lipschitz property of the loss function to relate the covering numbers of $\ell_{\mathcal{F}}$ and $\mathcal{F}$ as follows: 
\begin{lemma}\label{lemma: lipchitz_covering}
Let $\ell : \mathbb{Y} \times \mathbb{Y} \mapsto [0,H]$ be a loss function such that it satisfies:
$$\abs{\ell(y_1, y) - \ell(y_2,y)} \le L \cdot \abs{y_1 - y_2}.$$
Then, for any real valued function family $\mathcal{F}$, we have:
$$\mathcal{N}\left(\epsilon, \ell_{\mathcal{F}},m \right)\le \mathcal{N}\left(\frac{\epsilon}{L}, \mathcal{F}, m\right).$$
\end{lemma}
\begin{proof}
Let $S = \{(x_1, y_1) \ldots (x_m,y_m)\} \in \left(\mathbb{X}\times \mathbb{Y}\right)^{m}$, and let $g,h \in \mathcal{F}$ be two functions. We have:
\begin{equation*}
    \frac{1}{m}\sum_{i=1}^{m}\abs{\ell_{g}(x_i,y_i) - \ell_{h}(x_i,y_i)} 
    = \frac{1}{m}\sum_{i=1}^{m}\abs{\ell(y_i,g(x_i)) - \ell(y_i,h(x_i))}
    \le \frac{L}{m}\sum_{i=1}^{m}\abs{g(x_i)-h(x_i)}.
\end{equation*}
%
Hence, any $\frac{\epsilon}{L} $ cover for $\mathcal{F}_{\mid x_1^{m}}$ is an $\epsilon$ cover for $(\ell_{\mathcal{F}})_{\mid S}$. 
\end{proof}

Now we are ready for the proof of Lemma~\ref{lemma: bound_bad_prob2}. 

\begin{proof}[Proof of Lemma~\ref{lemma: bound_bad_prob2}]

We have:
\begin{align*}
    \mathbb{P}_{S\sim \mathbb{D}^{m}}& \left[\sup_{f\in \mathcal{F}}\abs{\hat{\err}_{S}(f) - \hat{\err}_{\mathbb{D}}(f)} \ge \eta\right]  \\
    &\le 2\cdot \mathbb{P}_{(S,\bar{S})\sim D^{m} \times D^{m}}[\sup_{f\in \mathcal{F}}\left(\abs{\hat{\err}_{S}(f) - \hat{\err}_{\bar{S}}(f)} \ge \eta/2\right)~ \text{(by Lemma~\ref{lemma: symmetrization})}\\
    &\le 2\cdot \sup_{(S,\bar{S}) \in (\mathbb{X}\times \mathbb{Y})^{2m}}\left(\mathbb{P}_{\sigma \sim U^m}\left[ \abs{\hat{\err}_{\sigma(S)} (f) - \hat{\err}_{\sigma(\bar{S})}(f)}\ge \eta/2 \right]\right)~ \text{(by Lemma~\ref{lemma: permutation})}\\
    &\le 2\cdot \mathcal{N}\left(\eta/8, \ell_{\mathcal{F}}, 2m\right) \cdot \max_{g \in \mathcal{G}} \left(\mathbb{P}_{\sigma \sim U^{m} }\left[\abs{\hat{\err}_{\sigma(S)} (g) - \hat{\err}_{\sigma(\bar{S})}(g)}\ge \eta/4 \right]\right)~\text{(by Lemma~\ref{lemma: quantization})} \\
    &\le 4\cdot \mathcal{N}\left(\eta/8, \ell_{\mathcal{F}}, 2m\right) \cdot exp\left(-\frac{m\cdot \eta^2}{64H}\right)~\text{(by Lemma~\ref{lemma: bernstein}).}
\end{align*}
\end{proof}

Finally, we arrive at the proof of Lemma~\ref{lemma: bound_bad_prob}.
\begin{proof}[Proof of Lemma~\ref{lemma: bound_bad_prob}]
Given that, $\big|\err_{\mathbb{D}, \epsilon}(f) - \err_{S, \epsilon}(f)\big| \le \eta\cdot \sqrt{\err_{\mathbb{D},  \epsilon}(f)}$, we claim:
\begin{equation*}
\err_{\mathbb{D},\epsilon}(f) \le (1+\alpha)\cdot \err_{S, \epsilon}(f) + \left(1+\frac{1}{\alpha}\right)\cdot \eta^2
\end{equation*}, and

\begin{equation*}
\err_{S,\epsilon}(f) \le \frac{2\alpha+1}{(1+\alpha)}\cdot \err_{\mathbb{D}, \epsilon}(f) + \left(1+\frac{1}{\alpha}\right)\cdot \eta^2
\end{equation*}

To show this, we consider the following two cases:

\begin{enumerate}
    \item If $\err_{\mathbb{D}, \epsilon}(f) \le \left(1+\frac{1}{\alpha}\right)^2\cdot \eta^2$, then we have $\big|\err_{\mathbb{D},\epsilon}(f) - \err_{S, \epsilon}(f) \big| \le \left(1+\frac{1}{\alpha}\right)\cdot \eta^2$. \item Otherwise, we have $\err_{\mathbb{D}, \epsilon}(f) > \left(1+\frac{1}{\alpha}\right)^2\cdot \eta^2$, and we get $\err_{\mathbb{D}, \epsilon}(f) \le \err_{S,\epsilon}(f) + \frac{\alpha}{1+\alpha}\cdot \err_{\mathbb{D}, \epsilon}(f)$, and $\err_{S, \epsilon}(f) \le \frac{2\alpha+1}{(\alpha+1)}\cdot \err_{\mathbb{D}, \epsilon}(f)$
\end{enumerate}
In either case, both the claims follow.

Lastly, the $\epsilon$ parameterised loss function $\ell_{\epsilon}(\cdot , \cdot )$ is $\frac{1}{\epsilon}$-Lipschitz in its first argument, from Lemma~\ref{lemma: lipchitz_covering}, we get that:
$$\mathcal{N}\left(\frac{\eta}{8}, \ell_{\mathcal{F}}, 2m \right) \le \mathcal{N}\left(\frac{\epsilon\cdot\eta}{8}, \mathcal{F}, 2m \right).$$
\end{proof}

We combine these results to present the proof of Theorem~\ref{thm:restricted}.

\begin{proof}[Proof of Theorem~\ref{thm:restricted}]
By Lemma~\ref{lemma: learner_to_alg}, it suffices to show that there exists a learning algorithm $\mathcal{L} : S^{m} \mapsto \mathcal{F}$ that outputs a function $\hat{f} : \mathbb{X}\mapsto \mathbb{Y} \in \mathcal{F}$ such that $\err_{\mathbb{D}, \epsilon}(\hat{f}) \leq 4\epsilon$. 
%
%
Recall that in the training phase of Algorithm~\ref{Alg: online_cost_with_pred_params}, we use an $\epsilon-$SEM algorithm $\mathcal{O}$ that returns a function $\hat{f}$ satisfying:
\begin{equation}
\label{eq:standard}
\err_{S,\epsilon}(\hat{f}) \le \inf_{\tilde{f}\in \mathcal{F}} \err_{S, \epsilon}(\tilde{f}) +\epsilon = \epsilon,
\end{equation}
where the last equality is because $\inf_{\tilde{f}\in \mathcal{F}} \err_{S, \epsilon}(\tilde{f}) = 0$ in the standard model.
So, we are left to bound $\err_{\mathbb{D}, \epsilon}(\hat{f})$ in terms of $\err_{S, \epsilon}(\hat{f})$, in particular, that $\err_{\mathbb{D}, \epsilon}(\hat{f}) \le 2\cdot \err_{S, \epsilon}(\hat{f})+\epsilon$, which would prove the theorem.

For this purpose, we employ Lemma~\ref{lemma: bound_bad_prob}. In this lemma, let us set $\alpha = 1$, and $\eta^2 = \epsilon$ and denote the event 
$$\sup_{f \in \mathcal{F}}\err_{\mathbb{D}, \epsilon}(f) \le 4\epsilon$$
as the ``good'' event; if this does not hold, we call it the ``bad'' event.

This leaves us to bound the probability of the bad event, which by Lemma~\ref{lemma: bound_bad_prob}, is at most $$4\cdot \mathcal{N}_{1}\left( \frac{\epsilon^{\frac{3}{2}}}{8}, \mathcal{F}, 2m \right) \cdot \exp\left(-\frac{m\cdot \epsilon}{64H}\right).$$ This quantity is at most $\delta$ when $m \ge C \cdot \left(\frac{H \cdot d\cdot \log \frac{1}{\epsilon} \cdot \log \frac{1}{\delta}}{\epsilon}\right)$ for a large constant $C$, thereby proving the theorem.
%
%
\end{proof}

Let us now move to the lower bound : Consider the input sequence $\Sigma = \tau_0, \tau_1, \ldots$ such that $\Off(0) =2, \Off(1) = 4$. Note that the log-cost at the two time-steps are 1 and 2 respectively. Let $\mathbb{X}$ be  set of $d$ distinct points (on the real line). Let $\cF$ be the set of all $2^d$ functions from $\cX$ to $\{0,1\}$. Clearly, the VC-dimension of $\cF$ is given by $d$. For every $f \in \cF$, we define a distribution $\dist_f$ over pairs $(x,y) \in \cX \times \{1,2\}$ as follows: $\dist_f$ is the uniform distribution over $A_f := \{(x,f(x)+1): x \in \cX\}$. Note that this is an instance of the standard setting, because for any distribution $\dist_f$, the corresponding function $f$ maps $x$ to $y$. 

Let $\cA$ be an algorithm for the LTS problem as above which has expected competitive ratio at most   $1+\varepsilon/4$ with probability at least $1-\delta$. 
Let $k$ be an upper bound on the sample complexity of $\cA$.
The algorithm $\cA$, after seeing $k$ samples, outputs a strategy. The strategy gives for each $x \in \cX$, a probability distribution over strategies~(i) and~(ii) as in the previous case. 

Now consider the following prediction problem $\cP$: we choose a function $f$ uniformly at random from $\cF$, and are given $k$ i.i.d. samples from $\dist_f$. We would like to predict a function $f' \in \cF$ which agrees with $f$ on at least $1-\varepsilon$ fraction of the points in $\cX$. 

\begin{lemma}
\label{lem:rest}
Suppose the algorithm $\cA$ has the above-mentioned properties. Then given $k$ i.i.d. samples from an instance of $\cP$, we can output the desired function $f'$ with probability at least $1-\delta$. 
\end{lemma}
\begin{proof}
Suppose the function $f$ gets chosen. We feed the $k$ i.i.d. samples from $\dist_f$ to $\cA$. The algorithm $\cA$ outputs a strategy $S$ which, for each $x$, gives a distribution $(q_x, 1-q_x)$ over strategies~(i) and (ii). 

Given this strategy $S$, we output the desired function $f'$ as follows. For every $x \in \cX$, if $q_1(x) \geq 1/2$, we set $f'(x)=1$, else we set it to 0. We claim that if $\cA$ has expected competitive ratio at most $1+\varepsilon$, then $f'$ agrees with $f$ on at least $\varepsilon$ fraction of points in $\cX$. 

Suppose not. Suppose $f(x) \neq f'(x)$ for some $x \in \cX$. If $f(x)=0$, then the cost of the optimal strategy here is 2, whereas the algorithm $\cA$ follows strategy~(ii) with probability at least $1/2$, and its expected cost is more than $2 \cdot \frac{1}{2} + 4 \cdot \frac{1}{2} = 3$. Similarly, if $f(x)=1$, optimal strategy pays 4. But algorithm $\cA$ places at least $1/2$ probability on strategy~(i). Therefore, its expected cost is more than  $\frac{1}{2} \cdot 4 + \frac{1}{2} \cdot 6 = 5.$ In either case, it pays at least 1.25 times the optimal cost. Since $f$ and $f'$ disagree on at least $\varepsilon$-fraction of the points, it follows that the expected competitive ratio of $\cA$ (when $x$ is chosen uniformly from $\cX$) is more than $1 + \varepsilon/4$, a contradiction. 

Since $\cA$ has competitive ratio at most $1+\varepsilon/4$ with probability at least $1-\delta$, the desired result follows. 
\end{proof}

Now, it is well known that if we want to find a function $f' \in \cF$ which matches with $f$ on more than $1-\varepsilon$ fraction of points in $\cX$ with probability at least $1-\delta$, we need to sample at least 
$\Omega \left( \frac{d}{\varepsilon} \ln \left( \frac{1}{\delta} \right) \right)$ points from $\dist_f$ (see Thm 5.3 in \cite{anthony1999learning}). This proves Theorem~\ref{thm:restlower}.

\subsection{Extension to the Agnostic Model}
In the agnostic model, we no longer assume a function $f \in \mathcal{F}$ that predicts the log-cost $y$ perfectly. It is possible that the true predictor is outside  $\mathcal{F}$, or in more difficult scenarios for any feature $x$, the behaviour of the log-cost $y$ may be entirely arbitrary. 

We first show that the loss function $\epsilon$-parameterized competitive error defined earlier is still a reasonable proxy for the competitive ratio. Specifically, we show that any algorithm that hopes to achieve a competitive ratio of $1+O(\epsilon)$ must use a prediction $\hat{f}\in \mathcal{F}$ whose error $\err_{\mathbb{D}, \epsilon}(f)$ is bounded by $O(\epsilon)$. We formally state this below:
%
\begin{lemma}\label{lem:losslowerbound}
Let $\mathcal{A}$ be an algorithm for \ltos that has access to a predictor $\hat{f}: \mathbb{X}\mapsto [0,H]$ for the log-cost $y$. Then, there exists a distribution $\mathbb{D}$ and a function $\hat{f}_{\mathcal{A}}$ with the property $\err_{\mathbb{D}, \epsilon}(\hat{f}_\mathcal{A}) = \epsilon$ such that 
$\ex_{(x,y)\sim \mathbb{D}}\left[\comp_{\mathcal{A}}(x,y)\right] \ge 1 + \frac{\epsilon}{2}$. 
\end{lemma}

\begin{proof}

Let the predicted log-cost be $\hat{y} = f_{\mathcal{A}}(x)$. Let $\phi(\hat{y})$ be the sum total of the costs of solutions bought by $\mathcal{A}$ till the optimal log-cost reaches
$\hat{y}$. Clearly $\phi(\hat{y}) \geq e^{\hat{y}}$. Since the algorithm $\mathcal{A}$ can be possibly randomized, let $e^{\hat{y}} \leq \phi(\hat{y}) \leq e^{\hat{y}+\epsilon}$ with probability $\alpha$ over the distribution chosen by $\mathcal{A}$.

We define the distribution $\dist$ as: $\cX$ is just the singleton set $\{x_0\}$ and $\cY = \{\hat{y}, \hat{y}\cdot\left(1 + {\epsilon}\right)\}$. The distribution $\dist$ assigns probability $1-{\epsilon}$ to $(x_0, \hat{y})$ and 
$\epsilon$ to $(x_0, \hat{y}\cdot\left(1 + {\epsilon}\right))$ (note that the optimal cost is $e^{\hat{y}}$ and $e^{\hat{y}\cdot\left(1 + {\epsilon}\right)}$ in these cases respectively). 
Note that $\dist$ and $f_{\mathcal{A}}$ satisfy $\err_{\mathbb{D}, \epsilon}(\hat{f}_\mathcal{A}) = \epsilon$.
The expected competitive ratio of $\cA$ is a least 


\begin{equation*}
    \comp 
    \geq (2\alpha + 1 - \alpha)\cdot {\epsilon} + \alpha\cdot(1-{\epsilon}) + (1-{\epsilon})\cdot (1-\alpha)\cdot (1+\epsilon)
    = 1 + \epsilon - (1-\alpha)\cdot \epsilon^2
    \geq 1+\frac{{\epsilon}}{2}.\qedhere
\end{equation*}    
%
\end{proof}

Unlike in the standard model, we no longer have that for any $\epsilon>0$, $\min_{f \in \mathcal{F}}\err_{\mathbb{D}, \epsilon}(f) = 0$. Therefore, we need to first quantify the performance of an {\em ideal} algorithm that uses predictors from $\mathcal{F}$.
\begin{definition}
Let $\chi (\epsilon) = \min_{f \in \mathcal{F}}\err_{\mathbb{D}, \epsilon}(f)$. Then, $\Delta_{\mathcal{F}}$ is the solution to the equation: $\epsilon = \chi(\epsilon)$.
\end{definition}

$\Delta_{\mathcal{F}}$ measures the best competitive ratio that we can hope to get when we use a predictor from $\mathcal{F}$.
Note that $\epsilon$ appears in two places in this definition, since the loss function in Definition~\ref{def: comp_loss} depends on $\epsilon$. We first show that this is a reasonable definition in that the solution to the equation is unique:

\begin{lemma}
\label{lem:unique}
For a given function family $\mathcal{F}$ and distribution $\mathbb{D}$, the value of $\Delta_{\mathcal{F}}$ is unique.
\end{lemma}
\begin{proof}[Proof of Lemma~\ref{lem:unique}]
Let $\chi (\epsilon) = \min_{f \in \mathcal{F}}\err_{\mathbb{D}, \epsilon}(f)$. Note that, $\chi (\epsilon)$ is non-increasing in $\epsilon$, and $\lim_{\epsilon \rightarrow 0} \chi(\epsilon) > 0$. Since $\ell_{\epsilon}(\cdot, \cdot) \le \frac{5}{\epsilon}-1$, we have $\chi(2) < 2$. Therefore, there must exist $\Delta_{\mathcal{F}} \in (0,2)$ such that:
$$\Delta_{\mathcal{F}} = \min_{f \in \mathcal{F}}\err_{\mathbb{D}, \Delta_{\mathcal{F}}}(f)$$
The uniqueness follows from the monotonicity of the function $\chi(.) $\qedhere
\end{proof}

We also give an algorithm that can approximate $\Delta_{\mathcal{F}}$ (Algorithm~\ref{Alg: calc_eps}).
\begin{algorithm}[ht]
\caption{Procedure to estimate $\Delta_{\mathcal{F}}$} 
\label{Alg: calc_eps}
\textbf{Input:}
 Sample Set $S$, and function family $\mathcal{F}$\\
 Let $\epsilon$ be an accuracy parameter given by the size of the sample set $S$.\\
 Choose $\varepsilon := \epsilon$\\
 Compute: $\hat{f}$ such that $\err_{S, \varepsilon}(\hat{f}) \le \min_{f\in \mathcal{F}} \err_{S, \varepsilon}(f) + \frac{\varepsilon}{3}$.\\
 \textbf{while $\varepsilon \le \err_{S, \varepsilon}(\hat{f})$}\\
 \hspace*{10pt} $\varepsilon \leftarrow 2\varepsilon$.\\
 \hspace*{10pt} Recompute $\hat{f}$ s.t. $\err_{S, \varepsilon}(\hat{f}) \le \min_{f\in \mathcal{F}} \err_{S, \varepsilon}(f) + \frac{\varepsilon}{3}$.\\
\textbf{Return } $\varepsilon$.
\end{algorithm} 


\begin{lemma}\label{lemma: calc_eps}
If $|S| \ge C\cdot \left(\frac{H\cdot d \log \frac{1}{\epsilon } \log \frac{1}{\delta}}{\epsilon}\right)$ for suitable constants $C > 0, \delta \le \frac{1}{2}$, and $\epsilon \le \Delta_{\mathcal{F}}$, then with probability at least $1-\delta$, we have $\frac{5}{36}\cdot \varepsilon \le \Delta_{F} \le \frac{17}{8}\varepsilon$, 
where $\varepsilon$ is as returned by Algorithm~\ref{Alg: calc_eps}.
\end{lemma}
\begin{proof}[Proof of Lemma~\ref{lemma: calc_eps}]
Let $\chi (\varepsilon) = \min_{f\in \mathcal{F}}\err_{\mathbb{D}, \varepsilon}(f)$ and $\lambda (\varepsilon) = \min_{f\in \mathcal{F}}\err_{S, \varepsilon}(f)$, where $S \sim \mathbb{D}^{m}$. For fixed $\hat{y}, y,$ we note that $\ell_{\epsilon}(y,\hat{y})$ can only decrease when $\epsilon$ increases. Therefore, both $\chi (\varepsilon)$ and $\lambda(\varepsilon)$ are non-increasing with $\varepsilon$.





From Lemma~\ref{lemma: bound_covering_number}, we have : $\mathcal{N}\left(\frac{\varepsilon^{\frac{3}{2}}}{8}, \mathcal{F}, 2m\right) \le \left(\frac{1}{\varepsilon}\right)^{O(d)}$. Noting that the size of the sample set $m$ exceeds $C\cdot \left(\frac{H\cdot d\cdot \log \frac{1}{\epsilon}\cdot \log \frac{1}{\delta}}{\epsilon}\right)$ for some large $C\ge 0$, we use Lemma~\ref{lemma: bound_bad_prob} with $\eta^2 = \frac{\varepsilon}{16}$ and $\alpha=1$ to claim that with probability $1-\delta$, we have for all $f\in \mathcal{F}$:
\begin{equation}\label{eq: relate_sample_error}
    \err_{S, \varepsilon}(f) \le \frac{3}{2}\cdot \err_{\mathbb{D}, \varepsilon}(f) + \frac{\varepsilon}{8}.
\end{equation}

and, 

\begin{equation}\label{eq: relate_exp_error}
    \err_{\mathbb{D}, \varepsilon}(f) \le 2\cdot \err_{S, \varepsilon}(f) + \frac{\varepsilon}{8}.
\end{equation}

Due to the breaking condition, we have $\varepsilon \ge \err_{S, \varepsilon}(\hat{f}) \ge \lambda (\varepsilon)$. Then, by Eq.~\eqref{eq: relate_exp_error}, we have:
\begin{equation*}
    \chi (\varepsilon) 
    \le 2\cdot \lambda(\varepsilon) + \frac{\varepsilon}{8}
    \le \frac{17}{8}\cdot \varepsilon.   
\end{equation*}
By monotonicity of $\chi (.)$,
\begin{equation}\label{eq: 2}
\chi \left(\frac{17}{8}\varepsilon\right)\le \chi \left(\varepsilon\right) \le \frac{17}{8}\cdot \varepsilon.
\end{equation}
Also, we have that $\frac{\varepsilon}{2} < \lambda (\frac{\varepsilon}{2}) + \frac{\varepsilon}{6}$. 
Let $f^{*} = \argmin_{f\in \mathcal{F}}\err_{\mathbb{D}, \frac{\varepsilon}{2}}(f)$, then using Eq.~\eqref{eq: relate_sample_error} on $f^{*}$, we get:
\begin{align*}
    \lambda \left(\frac{\varepsilon}{2}\right) &\le \err_{S, \frac{\varepsilon}{2}}(f^{*})\\
    &\le \frac{3}{2}\cdot \chi \left(\frac{\varepsilon}{2}\right) + \frac{\varepsilon}{8}\\
    \intertext{Hence, }
    \frac{5\varepsilon}{36} &\le \chi \left(\frac{\varepsilon}{2}\right)\le \chi \left(\frac{5\varepsilon}{36}\right).
\end{align*}

Combining the above with Eq.~\eqref{eq: 2}, we get $\frac{5}{36}\cdot \varepsilon \le \Delta_{F} \le \frac{17}{8}\varepsilon$.
\end{proof}

We are now ready to define our \ltos algorithm for the agnostic model. This algorithm is simply Algorithm~\ref{Alg: online_cost_with_pred_params} where the accuracy parameter $\epsilon$ is set to the value of $\varepsilon$ returned by Algorithm~\ref{Alg: calc_eps}.


\begin{theorem}\label{thm:general}
In the agnostic model for a function family $\mathcal{F}$, Algorithm~\ref{Alg: online_cost_with_pred_params} with accuracy parameter $\varepsilon$ from Algorithm~\ref{Alg: calc_eps} obtains a competitive ratio of $1+O\left(\Delta_{\mathcal{F}} \right)$ with probability at least $1-\delta$, when using $O\left(\frac{H\cdot d \log \left(\frac{1}{\Delta_{\mathcal{F}}}\right) \cdot \log \frac{1}{\delta}}{\Delta_{\mathcal{F}}}\right)$ samples, where $d=\pdim(\mathcal{F})$.
\end{theorem}


\begin{proof} 
From the breaking condition in Algorithm~\ref{Alg: calc_eps} and Lemma ~\ref{lemma: calc_eps}, we have that $\argmin_{f\in \mathcal{F}}\err_{S, \varepsilon}(f) \le \varepsilon \le \frac{36}{5}\cdot \Delta_{\mathcal{F}}$. Using the sample error minimization algorithm returns a function $\hat{f}$ such that 
\begin{align*}
\err_{S, \varepsilon}(\hat{f}) \leq \argmin_{f\in \mathcal{F}}\err_{S, \varepsilon}(f) + \varepsilon\le 2\varepsilon
\end{align*}
Finally, application of Lemma~\ref{lemma: bound_bad_prob} to bound $\err_{\mathbb{D}, \varepsilon}(\hat{f}) = O\left(\Delta_{\mathcal{F}}\right)$, followed by Lemma~\ref{lemma: learner_to_alg} gives the desired result.
\end{proof}

We also lower bound the sample complexity of a \ltos algorithm:
\begin{restatable}{theorem}{lb1}
\label{thm:lowergen}
Any \ltos algorithm  that is $\epsilon$-efficient with probability at least $1-\delta$ must query $\Omega\left(\frac{\log \frac{1}{\delta}}{\epsilon^2}\right)$ samples.
\end{restatable}

We begin with the agnostic case. We describe a class of distributions $\dist_p$ on  pairs $(x,y)$, where $p$ is a parameter in $(0,1)$. Recall that $y$ represents $\log_2 z$, where $z$ is the actual optimal cost of the offline-instance.  The distribution $\dist_p$ consists of two pairs: $(1, 1)$ with probability $p$, and $(0,2)$ with probability $1-p$. Note that the projection of $\dist_p$ on the first coordinate is a Bernoulli random variable with probability of 1 being $p$. For the sake of concreteness, the input sequence $\Sigma = \tau_0, \tau_1, \ldots, $ is such that $\Off(0)=2, \Off(1)=4$. The distribution $\dist_p$ ensures that the stopping time parameter $T=0$ with probability $p$, and $T=1$ with probability $1-p.$ It follows that any online algorithm has only one decision to make: whether to buy the solution for $\cI_0$. 

Let $\Ast_p$ be the algorithm which achieves the minimum competitive ratio when the input distribution is $\dist_p$, and let  $\rho^\star_p$ be the expected competitive ratio of this algorithm.
There are only two strategies for any algorithm: (i) buy optimal solution for $\cI_0$, and if needed buy the solution for $\cI_1$, or (ii) buy the optimal solution for $\cI_1$ at the beginning. 
The following result determines the value of $\rho^\star_p$. 

\begin{lemma}
\label{lem:comp}
If $p = \frac{1}{3} + \varepsilon$ for some $\varepsilon \geq 0$, then 
$\rho^\star_p = \frac{4}{3} - \frac{\varepsilon}{2},$ and strategy~(i) is optimal here. In case $p = \frac{1}{3} - \varepsilon$ for some $\varepsilon \geq 0$, then $\rho^\star_p = \frac{4}{3} - \varepsilon,$ and strategy~(ii) is optimal. 
\end{lemma}

\begin{proof}

For strategy~(i), the cost of the algorithm is $2$ with probability $p$ and $6$ with probability $1-p$. Therefore its expected competitive ratio is 
$$ p \cdot 1 + \frac{6}{4} \cdot (1-p) = \frac{3}{2} - \frac{p}{2}.$$

For strategy~(ii), the cost of the algorithm is always 4. Therefore, its expected competitive ratio is 
$$ p \cdot 2 + 1 \cdot (1-p) = p + 1. $$

It follows that strategy~(i) is optimal when $p \geq 1/3$, whereas strategy~(ii) is optimal when $p \leq 1/3$. 
\end{proof}

We are now ready to prove Theorem~\ref{thm:lowergen}. Let $\cA$ be an algorithm for LTS which is $\varepsilon/4$-efficient with probability at least $1-\delta$. Further, let $k$ be an upper bound on the sample complexity of $\cA$. Given $k$ samples from a distribution $\dist_p$, the algorithm outputs a strategy which is a probability distribution on strategies~(i) and~(ii). We use this algorithm $\cA$ to solve the following prediction problem~$\cP$: $X$ is a random variable uniformly distributed over $\{\frac{1}{3} - \varepsilon, 
\frac{1}{3} + \varepsilon \}.$ Given i.i.d. samples from from $0$-$1$ Bernoulli random variable $T$ with probability of 1 being $X$, we would like to predict the value of $X$. 


\begin{lemma}
\label{lem:pred}
Let $\cA$ be an algorithm for LTS which is $\varepsilon/4$-efficient with probability at least $1-\delta$.
Then, there is an algorithm that predicts the value of $X$ with probability at least $1-\delta$ using $k$ i.i.d. samples from $T$. 
\end{lemma}
\begin{proof}
Let $t_1, \ldots, t_k$ be i.i.d. samples from $T$. We give $k$ samples $(x_1, y_1), \ldots, (x_k, y_k)$ to $\cA$ as follows: for each $i = 1, \ldots, k$, if $t_i=0$, we set $(x_i, y_i)$ to $(0,2)$; else we set it to $(1,1)$. Observe that the samples given to $\cA$ are $k$ i.i.d. from the distribution $\dist_X$. 

Based on these samples, $\cA$ puts probability $q_1$ on strategy~(i) (and $1-q_1$ on strategy~(ii)). If $q_1 > 1/2$, we predict $X = \frac{1}{3} + \varepsilon$, else we predict $X = \frac{1}{3} - \varepsilon$. 

We claim that this prediction strategy predicts $X$ correctly with probability at least $1-\delta$. To see this,  assume that $\cA$ is $\varepsilon/4$-efficient (which happens with probability at least $1-\delta$). 

First consider the case when $X = \frac{1}{3} + \varepsilon$. In this case, Lemma~\ref{lem:comp} shows that the expected competitive ratio of $\cA$ is at most $\frac{4}{3} - \frac{\varepsilon}{4}.$ 
As in the proof of Lemma~\ref{lem:comp}, the expected competitive ratio of $\cA$ is 
$$ q_1 \left(\frac{3}{2} - \frac{X}{2} \right) + (1-q_1) ( X+1). $$
We argue that $q_1 \geq 1/2$. Suppose not. 
Since $X > 1/3$, $\frac{3}{2} - \frac{X}{2} \leq X+1$. Therefore, the above is at least (using $q_1 \leq 1/2$ and $X = 1/3 + \varepsilon$) 
$$ \frac{1}{2} \left(\frac{3}{2} - \frac{X}{2}\right) + \frac{1}{2} ( X+1) > 4/3,$$
which is a contradiction. Therefore $q_1 >  1/2$. 

Now consider the case when $X = 1/3 - \varepsilon$. Again Lemma~\ref{lem:comp} shows that the expected competitive ratio of $\cA$ is at most $\frac{4}{3} - \frac{3\varepsilon}{4}$. 
It is easy to check that if $q_1 \geq 1/2$, then the expected competitive ratio of $\cA$ is at least 
$$ \frac{4}{3} - \frac{\varepsilon}{4}, $$ which is a contradiction. Therefore, $q_1 < 1/2$. This proves the desired result. 
\end{proof}

It is well known that in order to predict $X$ with probability at least $1-\delta$, we need $\Omega \left( \frac{1}{\varepsilon^2} \ln \left( \frac{1}{\delta} \right) \right)$ samples. This proves Theorem~\ref{thm:lowergen}.

\subsection{Robustness of Algorithm~\ref{Alg: online_cost_with_pred_params}}


So far, we have established the competitive ratio of Algorithm~\ref{Alg: online_cost_with_pred_params} in the PAC model. Now, we show the robustness of this algorithm, i.e., bound its competitive ratio for {\em any} input. 
Even for adversarial inputs, we show that this algorithm has a competitive ratio of $O(1/\epsilon)$, which matches the robustness guarantees in Theorem~\ref{thm:robust-consistent} for the \pdbl algorithm. 
\begin{theorem}
\label{thm:robust}
Algorithm~\ref{Alg: online_cost_with_pred_params} is $5(1 + \frac{1}{\epsilon}) = O\left(\frac{1}{\epsilon}\right)$-robust.
\end{theorem}

\begin{proof}[Proof of Theorem~\ref{thm:robust}]
We show this theorem by using the following lemma:
\begin{lemma}
\label{lem:ltos-robust}
Let $\mathcal{A}$ denote Algorithm~\ref{Alg: online_cost_with_pred_params}. Then 
\begin{equation*}
    \comp_{\mathcal{A}}(x,y) = \begin{cases} 4 \text{  when } y\leq \ln \frac{5}{\epsilon}-\hat{y} \text{ or }y > \hat{y}+\ln\left(1+\frac{\epsilon}{5}\right)\\
    (1+\epsilon)\cdot e^{\hat{y}-y}\text{ otherwise}\\
    \end{cases}
\end{equation*}
\end{lemma}
\begin{proof}
The proof follows from Lemma~\ref{lemma: graceful_degradation} by noting that $e^{y}=\Off(T), e^{\hat{y}} = \Off(\hat{T})$, and $\Off(t_1) = \frac{\epsilon}{5}\cdot e^{\hat{y}}, \Off(t_2) = (1+\frac{\epsilon}{5})\cdot e^{\hat{y}}.$
\end{proof}

Theorem~\ref{thm:robust} now follows by noting that the worst case is when $y$ just exceeds $t_1$, i.e., when $y= \ln \frac{5}{\epsilon} - \hat{y}$.
\end{proof}

\section{Inadequacy of Traditional Loss Functions}
\label{sec:traditional}

In this section, we motivate the use of asymmetric loss function (Definition~\ref{def: comp_loss}) by showing that an algorithm which uses predictions from a learner minimizing a symmetric loss function, such as absolute loss or squared loss, would have a large competitive ratio. The intuition is that if we  err on either side of the true value of $T$ by the same amount, the competitive ratio in the two cases does not scale in the same manner. To formalize this intuition, we define a class of distributions $\dist_\Delta$, parameterized by $\Delta> 0$, which are symmetric around a real $c$; more concretely this distribution places equal weight on $\{c-\Delta, c+\Delta\}$. Any algorithm relying on a symmetric loss function would always predict $c$. In such a case, the online algorithm $\cA$ has no new information. However, if $\Delta$ is large, an offline algorithm is better off buying the solution till $c-\Delta$ first, whereas if $\Delta$ is small, it should buy the solution for $c+\Delta$ in the first step. An algorithm which relies only on $c$ would err in one of these two cases. This idea is formalized in Lemma~\ref{lem:losssym}. Our second result (Lemma~\ref{lem:lossabs}) shows that predicting log-loss within an additive $\epsilon$ factor may result in a $1 + \Omega\left(\sqrt{\epsilon}\right)$ expected competitive ratio. This further bolsters the case for the loss function  as in Definition~\ref{def: comp_loss}. 

\begin{lemma}
\label{lem:losssym}
Let $\mathcal{A}$ be an algorithm that uses predictions made by a learner that minimizes symmetric error. Then, one of the following statements is true:
\begin{enumerate}
    \item $\ex_{(x,y)\sim \dist_\Delta}\left[\comp_{\mathcal{A}}(x,y)\right]$ is $\Omega(e^{\Delta})$ when $\Delta \ge 4$.
    \item $\ex_{(x,y)\sim \dist_\Delta}\left[\comp_{\mathcal{A}}(x,y)\right] \ge 1+\Omega(1)$, when $\Delta = \epsilon,$ with $\epsilon$ being an arbitrarily small positive real number. 
\end{enumerate}
\end{lemma}
\begin{proof}

We  define a family of distributions $\dist_\Delta$, parameterized by $\Delta$, $0 \leq \Delta \leq c$, where $c$ is a large enough constant, as follows: 
\begin{definition}
Let $\cX$ denote the singleton set $\{x_0\}$ and $\cY$ denote $\{c-\Delta, c + \Delta\}$. 
The distribution $\dist_\Delta$ on $\mathbb{X}\times \mathbb{Y}$ assigns probability $\frac{1}{2}$ to both $(x_0, c- \Delta)$ and $(x_0, c+\Delta)$. 
\end{definition} 

Ideally, we would want an algorithm $\cA$ to satisfy $\ex_{(x,y)\sim \dist_\Delta}\left[\comp_{\mathcal{A}}(x,y)\right] \rightarrow 0$ when $\Delta \rightarrow 0$, while still maintaining a worst-case result like $\ex_{(x,y)\sim \dist_\Delta}\left[\comp_{\mathcal{A}}(x,y)\right] \le O(1)$. The following construction shows that using a symmetric loss function would not be helpful.
Suppose we use a learner that outputs the function which minimizes a symmetric loss function. Then given samples from $\dist_\Delta$, such a learner will always yield $\hat{y} = c$ as the prediction. 


Since the feature is fixed, the behavior of the algorithm is independent of the feature and hence, it only needs to decide on a list of solutions that it will progressively buy. Let $\tau$ be the cost of the first solution bought by $\mathcal{A}$ that lies inside the interval $[e^{c-\Delta},e^{c+\Delta}]$ where $\hat{y}=c$ is the predicted log-cost that has been supplied to the algorithm.

There are two possible cases for $\tau$:
\begin{itemize}
    \item[(i)]  $\tau \ge e^{c}$: With probability $\frac{1}{2}$, the competitive ratio is $\frac{e^{c}}{e^{c-\Delta}} = e^{\Delta}$. Hence, $\ex\left[\comp_{\mathcal{A}}(x,y)\right] \ge \frac{1+e^{\Delta}}{2}$. Observe that if $\Delta \geq 4$, then  $\ex\left[\comp_{\mathcal{A}}(x,y)\right]$ is $\Omega(e^{\Delta})$, and hence unbounded. 
    \item[(ii)] $\tau < e^{c}$: With probability $\frac{1}{2}$, $y = c+\Delta$, in which case the competitive ratio is $\frac{e^{c}+e^{c+\Delta}}{e^{c+\Delta}} = 1 + e^{-\Delta}$. Therefore, 
    $\ex\left[\comp_{\mathcal{A}}(x,y)\right]$ is $1 + \frac{e^{-\Delta}}{2} = 1 + \Omega(1)$, even when $\Delta$ is an arbitrarily small positive $\epsilon$.\qedhere
\end{itemize}
\end{proof}


It is worth noting that if we use the loss function as in Definition~\ref{def: comp_loss}, then Algorithm~\ref{Alg: online_cost_with_pred_params}  has expected competitive ratio
$1+O(\epsilon)$ when $\Delta \le \epsilon$. Further, this algorithm defaults to \dbl when $\Delta = \Omega(1)$, and hence has bounded competitive ratio in this case. 

We also show that any algorithm which relies on a predictor of log-cost which has an $\epsilon$ bound on the absolute loss must incur $1 + \Omega(\sqrt{\epsilon})$ expected competitive ratio. Comparing this result with Theorem~\ref{thm:general} shows that our loss function defined as in Definition~\ref{def: comp_loss} gives better competitive ratio guarantees.


\begin{lemma}
\label{lem:lossabs}
 Let $\mathcal{A}$ be a learning-augmented algorithm for \OS, that has access to a predictor $P: \mathbb{X}\mapsto [0,H]$ that predicts the log-cost $y$. Moreover, the only guarantee on $P$ is that $\ex_{(x,y)\sim \mathbb{D}}\left[\abs{P(x) - y}\right] \le \epsilon$. Then there is a distribution $\dist$ and a predictor $P$ such that $\ex_{(x,y)\sim \mathbb{D}}\left[\comp_{\mathcal{A}}(x,y)\right] \ge 1 + \frac{\sqrt{\epsilon}}{2}$. 

\end{lemma}

\begin{proof}

Fix an algorithm $\cA$.  
Given the prediction $\hat{y} = 1$,  the algorithm outputs a (randomized) strategy for buying optimal solutions at several time steps. 
Let $\phi$ be the sum total of the costs of the solutions bought by the algorithm before the cost of the optimal solution reaches $e$.
Clearly $\phi \geq e$.
Let $\alpha$ denote the probability that $\phi \in [e, e^{1+ \sqrt{\epsilon}}]$, where the 
probability is over the distribution chosen by $\cA$. 

We define the distribution $\dist$ as follows: $\cX$ is just the singleton set $\{x_0\}$ and $\cY = \{1, 1 + \sqrt{\epsilon}\}$. The distribution $\dist$ assigns probability $1-\sqrt{\epsilon}$ to $(x_0, 1)$ and 
$\sqrt{\epsilon}$ to $(x_0, 1+ \sqrt{\epsilon})$ (note that the optimal cost is $e$ and $e^{1 + \sqrt{\epsilon}}$ in these cases respectively). Note that $\ex_\dist[y]$ is $1 + \epsilon$. Consider the predictor $P$ which outputs the prediction $\hat{y} = 1$, and therefore satisfies the condition $\ex_{(x,y)\sim \mathbb{D}}\left[\abs{P(x) - y}\right] \le \epsilon$. The expected competitive ratio of $\cA$ is a least 
\begin{align*}
    (1- \sqrt{\epsilon}) \left[ \alpha \cdot 1 + (1- \alpha) \cdot e^{\sqrt{\epsilon}} \right] + \\
    \quad \quad \sqrt{\epsilon} \left[ \alpha \frac{e + e^{1+\sqrt{\epsilon}}}{e^{1 + \sqrt{\epsilon}}}  + (1- \alpha) \cdot 1 \right] 
\end{align*}

Approximating $e^{\sqrt{\epsilon}}$ by $1 + \sqrt{\epsilon},$ the above expression simplifies to $$ 1 + \sqrt{\epsilon} - \epsilon \geq 1 + \frac{\sqrt{\epsilon}}{2}.$$
%

\end{proof}

\section{Conclusion, Limitations, and Future Work}
\label{sec:conclusion}
In this paper, we studied the role of regression in making predictions for learning-augmented online algorithms. In particular, we used the \OS framework that includes a variety of online problems such as ski rental and its generalizations, online scheduling, online bin packing, etc. and showed that by using a carefully crafted loss function, we can obtain predictions that yield near-optimal algorithms for this problem. One assumption that holds for the above problems, but not for other problems such as online matching, is the composability of solutions, i.e., that the union of two feasible solutions is also a feasible solution. Extending our work to such ``packing'' problems is an interesting direction for future research. Another interesting direction would be to give a general recipe for converting competitive ratios to loss functions, minimizing which over a collection of training samples generates better \ml predictions for online problems.

\section{Acknowledgements}
\label{sec: acknowledge}
This research was partially funded by the Indo-US Virtual Networked Joint Center project No. IUSSTF/JC-017/2017. Keerti Anand and Debmalya Panigrahi were supported in part by NSF Awards CCF-1955703, CCF-1750140 (CAREER), and ARO Award W911NF2110230. Rong Ge was also supported in part by NSF Awards DMS-2031849, CCF-1704656, CCF-1845171 (CAREER), CCF-1934964 (TRIPODS), a Sloan Research Fellowship, and a Google Faculty Research Award.

\clearpage
\bibliographystyle{unsrt}
\bibliography{references}

\end{document}